\documentclass{article} % For LaTeX2e
\usepackage{url}
\PassOptionsToPackage{table}{xcolor}
\usepackage{iclr2025_conference,times}

\usepackage{dsfont}
\usepackage{fontawesome}
\usepackage{wrapfig}
\usepackage{algorithm}
\usepackage{algorithmic}
\usepackage{multirow}
\usepackage{graphicx}
\usepackage{booktabs}
\usepackage{caption}
\usepackage{adjustbox}
\usepackage{pifont}
\usepackage{epstopdf}
\usepackage{hyperref}
\hypersetup{  
    colorlinks=true,  
    citecolor=cyan,
}

\usepackage{amsmath,amsfonts,bm}

\def\eqref#1{equation~\ref{#1}}
\def\1{\bm{1}}

\DeclareMathAlphabet{\mathsfit}{\encodingdefault}{\sfdefault}{m}{sl}
\SetMathAlphabet{\mathsfit}{bold}{\encodingdefault}{\sfdefault}{bx}{n}

\title{O-Edit: Orthogonal Subspace Editing for Language Model Sequential Editing}

\author{
    Yuchen Cai, Ding Cao \\ % 作者名
    University of Science and Technology of China \\
    State Key Laboratory of Cognitive Intelligence \\
    \texttt{\{caiyuchen, caoding\}@mail.ustc.edu.cn} \\}

 \iclrfinalcopy % Uncomment for camera-ready version, but NOT for submission.
\begin{document}

\maketitle

\begin{abstract}
Large language models (LLMs) acquire knowledge during pre-training, but over time, this knowledge may become incorrect or outdated, necessitating updates after training. Knowledge editing techniques address this issue without the need for costly re-training. However, most existing methods are designed for single edits, and as the number of edits increases, they often cause a decline in the model's overall performance, posing significant challenges for sequential editing. To overcome this, we propose Orthogonal Subspace Editing, O-Edit. This algorithm orthogonalizes the direction of each knowledge update, minimizing interference between successive updates and reducing the impact of new updates on unrelated knowledge. Our approach does not require replaying previously edited data and processes each edit knowledge on time. It can perform thousands of edits on mainstream LLMs, achieving an average performance improvement that is 4.2 times better than existing methods while effectively preserving the model's performance on downstream tasks, all with minimal additional parameter overhead.
\end{abstract}

\section{Introduction}
Large language models (LLMs) are trained on vast amounts of textual data, enabling them to store extensive knowledge about various aspects of the human world, sparking the potential for general artificial intelligence. However, LLMs face significant challenges, including the propagation of inaccurate or outdated knowledge, as well as the generation of bias or harmful content \citep{cai2024locatingmitigatinggenderbias, chen2024largelanguagemodelbias, wang2024detoxifyinglargelanguagemodels}. Given the substantial computational costs of re-training LLMs to address these issues, there has been growing interest in model editing techniques \citep{yao2023editinglargelanguagemodels, wang2023knowledgeeditinglargelanguage}, which aim to update specific content within the model while minimizing computational costs. Existing model editing methods can be categorized into two main types: parameter-modifying methods that directly alter a small subset of model parameters \citep{dai2022knowledgeneuronspretrainedtransformers, meng2023locatingeditingfactualassociations, meng2023masseditingmemorytransformer,hu2024knowledgesuperpositionunveilingfailures,hu2024wilkewiselayerknowledgeeditor, gupta2024rebuildingromeresolving}, and parameter-preserving methods that without changing the model parameters \citep{wang2024wiserethinkingknowledgememory, cai2024editingknowledgerepresentationlanguage, zheng2023editfactualknowledgeincontext}. In this paper, we focus on parameter-modifying editing methods.

Most existing research focuses on editing models a single time \citep{han-etal-2023-improving, zhang2024dafnetdynamicauxiliaryfusion}. However, as real-world knowledge continuously evolves, models will need to be updated repeatedly to remain accurate. This shift has led to the concept of sequential model editing \citep{ma2024perturbationrestrainedsequentialmodelediting, hu2024wilkewiselayerknowledgeeditor, huang2023transformerpatchermistakeworthneuron}, which involves performing multiple knowledge edits to progressively update the model as new knowledge needs to be incorporated. Currently, sequential editing is often achieved through multiple iterations of single edits. Recent studies have shown that as the number of edits increases, the success rate of edits significantly declines and impairs the model's general capabilities, such as reasoning and contextual understanding, thereby limiting the scalability of model editing \citep{gu2024modeleditingharmsgeneral, gupta2024rebuildingromeresolving}. This challenge is akin to adding new floors to an existing building—each addition risks compromising the overall stability. While some research has analyzed the bottlenecks of sequential editing from a theoretical perspective \citep{ma2024perturbationrestrainedsequentialmodelediting, hu2024knowledgesuperpositionunveilingfailures}, there is still no effective solution to these problems. These unresolved issues pose significant obstacles to the practical implementation of model editing\footnote{For more details on related work, please refer to Appendix \ref{related work}.}.

\begin{wrapfigure}{r}{0.3\textwidth}
\vspace{-0.5cm}
    \centering
    \includegraphics[width=\linewidth]{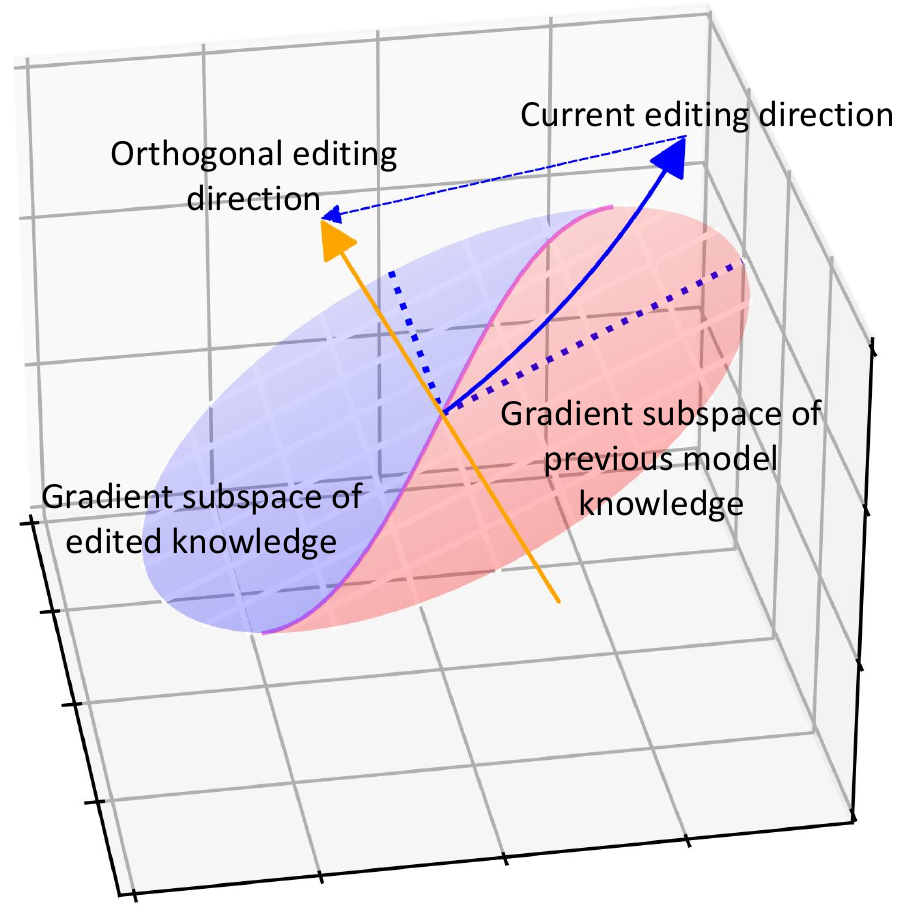}
    \caption{O-Edit constrains the direction of each update to lie within an orthogonal subspace.}
    \label{fig1}
 \vspace{-0.5cm}
\end{wrapfigure}

To address the scalability issue of sequential editing, this paper introduces Orthogonal Subspace Editing (O-Edit), a simple yet effective method for sequentially editing language models. Our key insight is based on the observation that existing editing methods primarily perform updates within specific low-rank subspaces. Based on this premise, we assume that both the update directions from previous editing tasks and the directions of updates to the model’s implicit knowledge can be captured. Therefore, for the current editing knowledge, the direction of parameter updates should be chosen to minimize the impact on these prior update directions. O-Edit accomplishes this by projecting the update direction of the current knowledge into an orthogonal subspace, ensuring that the neural network's output for previous knowledge remains unchanged while the projected direction remains effective for the current edit. To enhance O-Edit, we introduce O-Edit+, a post-processing method designed to ensure complete orthogonality between subspaces. We validate the effectiveness of our methods by utilizing two knowledge editing datasets and four downstream task datasets. Furthermore, our analysis, conducted from both experimental and theoretical perspectives, clearly demonstrates that strong orthogonality between each update matrix is crucial for enabling sequential editing. Figure \ref{fig1} illustrates how our methods adjust the update direction for each piece of knowledge.

Our method offers four key advantages: (1) \textbf{Efficiency}: It requires minimal additional parameters while enabling hundreds or even thousands of sequential edits. (2) \textbf{Privacy}: There is no requirement to store the edited data itself, ensuring privacy during updates. (3) \textbf{Timeliness}: Our method allows for the immediate application of each edit, making it more practical. (4) \textbf{Flexibility}: Our method is compatible with existing sequential editing techniques, allowing for easy integration and adaptability to various scenarios.

Our main contributions are as follows: \ding{172} We introduce O-Edit and O-Edit+, two simple and efficient methods for sequential editing in large language models (LLMs) that can handle thousands of edits in orthogonal subspaces, effectively addressing the performance degradation issue encountered by existing approaches during multiple edits. \ding{173} Our methods significantly preserve model performance on downstream tasks, demonstrating their scalability and practicality even after numerous sequential edits in real-world continuous model update scenarios. \ding{174} We show that the orthogonality between knowledge is essential for supporting sequential editing, providing a viable research direction for this task.

\section{PRELIMINARIES}
\label{2}

    In this section, we introduce sequential model editing. Subsequently, in Section \ref{3}, we discuss two prominent knowledge editing techniques, ROME \citep{meng2023locatingeditingfactualassociations} and MEMIT \citep{meng2023masseditingmemorytransformer}, and extend them into the sequential editing method O-Edit. Finally, in Section \ref{4}, we further refine O-Edit by presenting O-Edit+, a more straightforward and effective approach for orthogonal sequential model editing.

 We focus on the challenge of sequential model editing (SME) \citep{wang2024wiserethinkingknowledgememory, ma2024perturbationrestrainedsequentialmodelediting}, which aims to enable large language models (LLMs) to undergo extensive sequential modifications, potentially involving hundreds or thousands of edits. The primary objective is to ensure that the model's outputs align with human expectations across target queries, while simultaneously preserving the LLM's pre-existing knowledge and capabilities. Let \( f_{\Theta}: \mathbb{X} \rightarrow \mathbb{Y} \), parameterized by \( \Theta \), denote a model function that maps an input \( \mathbf{x} \) to its corresponding prediction \( f_{\Theta}(\mathbf{x}) \). The initial model, \( f_{\Theta_0} \), is pre-trained on a large dataset \( D_{\text{train}} \). When the LLM exhibits inaccuracies or requires updates, model editing becomes necessary, using a dynamic, time-evolving dataset \( \mathcal{D}_{\text{edit}} = \{(\mathcal{X}_e, \mathcal{Y}_e) \mid (x_1, y_1), \ldots, (x_T, y_T)\} \). At each time step \( T \), a model editor (ME) applies the \( T \)-th edit, updating the previous model \( f_{\Theta_{T-1}} \) to produce a new model \( f_{\Theta_T} \), following the equation:

\begin{equation}
f_{\Theta_T} = \text{ME}(f_{\Theta_{T-1}}, \mathbf{x}_T, y_T), \quad \text{s.t.} \quad f_{\Theta_T}(\mathbf{x}) = 
\begin{cases} 
y_T & \text{if } \mathbf{x} \in \mathcal{X}_e, \\
f_{\Theta_0}(\mathbf{x}) & \text{if } \mathbf{x} \notin \mathcal{X}_e.
\end{cases}
\label{eq:1}
\end{equation}

Eqn. \ref{eq:1} indicates that after model editing, the LLM should correctly predict the current edit with \( f_{\Theta_T}(\mathbf{x}_T) = y_T \), while preserving previous edits \((\mathbf{x}_{<T}, y_{<T}) \in \mathcal{D}_{\text{edit}}\) are inaccessible to the editor, the model is still able to retain this edit. Additionally, the model should maintain the performance of the original model \( f_{\Theta_0} \) on data outside the editing scope, \( \mathbf{x} \notin \mathcal{X}_e \), particularly with respect to the general training corpus \( D_{\text{train}} \).

\begin{figure}
\centering
\vspace{-1cm}
\includegraphics[width=0.8\textwidth]{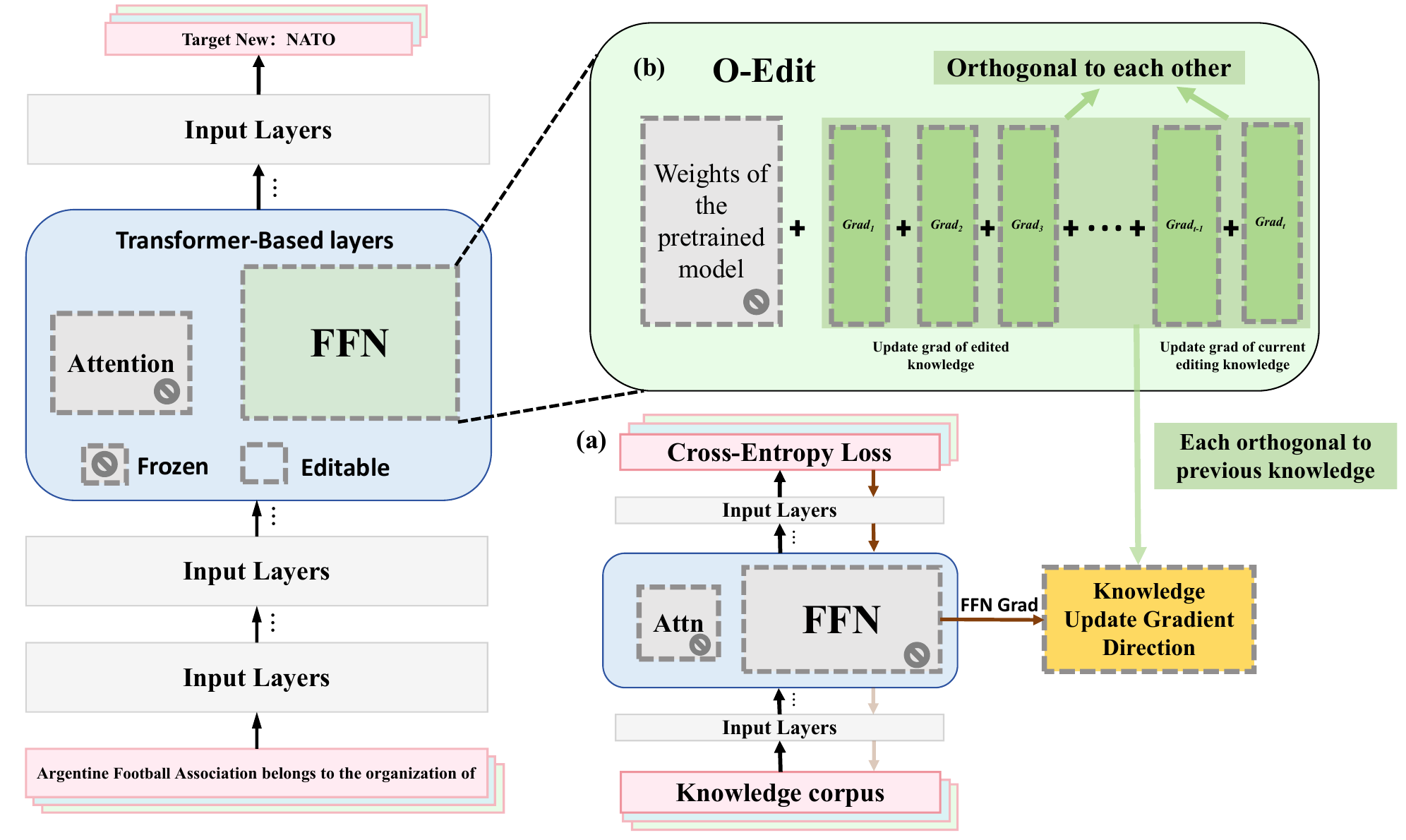}
\caption{The framework of O-Edit for sequential language model editing. \textbf{(a)} First, we compute gradients on a large amount of textual data without updating the model parameters. This step provides the gradient information necessary for updating model's implicit knowledge. \textbf{(b)} Next, we impose constraints on the update directions for each piece of edited knowledge, ensuring these directions are orthogonal to each other as well as to the directions of the model's implicit knowledge.}
\vspace{-0.2cm}
\label{fig2}
\end{figure}

\section{\textbf{O-EDIT}: SEQUENTIAL EDITING WITH GRADIENT PROJECTION MEMORY}
\label{3}

In this section, we introduce O-Edit, as illustrated in Figure \ref{fig2}. First, we discuss two key-value memory-based knowledge editing methods, ROME and MEMIT, followed by our optimization method, which incrementally edits new knowledge in orthogonal subspaces, while preserving previously edited knowledge.

\subsection{Key-Value Memory Editing Method}
In their study, \citep{meng2023locatingeditingfactualassociations} employed causal mediation analysis to identify that feed-forward neural networks (FFNs) play a crucial role in retaining factual knowledge. The FFN is decomposed into two matrices, represented as follows:
\begin{equation}
    FFN^{l}(x) = W^{l}_{proj} \cdot \sigma( W^{l}_{fc} \cdot \gamma (a^l + h^{l-1}))
\end{equation}
Here, \( a^l \in \mathbb{R}^{d} \) represents the output of the attention module at the \( l \)-th layer, and \( h^{l-1} \in \mathbb{R}^{d} \) denotes the output of the previous layer. The matrices \( W_{fc}^l \in \mathbb{R}^{d_m \times d} \) and \( W_{proj}^l \in \mathbb{R}^{d \times d_m} \) serve as the parameter matrices for the FFN at the \( l \)-th layer. Here, \( d_m \) is the dimension of the intermediate hidden state, \( \sigma \) denotes the activation function, and normalizing nonlinearity \( \gamma \).

Building on the key-value memory theory introduced in \citep{geva2021transformerfeedforwardlayerskeyvalue, geva2022transformerfeedforwardlayersbuild}, the matrix \( W^{l}_{fc} \) is responsible for identifying input patterns, which leads to the generation of the key vector $k \in \mathbb{R}^d_m$. In contrast, \( W^{l}_{proj} \) retrieves the corresponding value vector $v \in \mathbb{R}^d$. This establishes \( W^{l}_{proj} \) as a linear key-value memory system, where the set of key vectors \( K = \{k_1, k_2, \ldots\} \) is associated with the corresponding set of value vectors \( V = \{v_1, v_2, \ldots\} \). The relationship between the keys and values can be succinctly expressed as \( WK = V \), thereby completing the transformation process.

\cite{meng2023locatingeditingfactualassociations} propose \textbf{ROME}, in which new knowledge is represented as a key-value pair \((k_*, v_*)\) and is integrated into the model by addressing the following constrained least squares problem:

\begin{equation}
\label{eq5}
\min \|\widetilde{W}K - V\|_2 \quad \text{subject to} \quad \widetilde{W}{k_*} = v_*, \quad \text{with} \quad \widetilde{W} = W + \Lambda(C^{-1} k_*)^T.
\end{equation}

Here, \( \Delta W = \Lambda(C^{-1} k_*)^T \), \( k_* \) represents the query associated with the knowledge to be edited, such as \( x = \textit{“The president of the US is”} \), where \( k_* \) corresponds to the hidden state of the last token (index \( i \)) of the subject (e.g., \textit{“US”}). The key vector \( k_* \) is defined as:

\vspace{-0.4cm}
\begin{equation}
\label{eqk}
k_* = \frac{1}{N} \sum_{j=1}^N k(s_j + x), \quad \text{where } k(x) = \sigma\left( W_{fc}^{l}  \gamma \left(a^{l}_{[x],i} + h_{[x],i}^{l-1} \right) \right),
\end{equation}
\vspace{-0.4cm}

with \( s_j \) representing prefix texts for robustness. The value vector \( v_* \) denotes the edited knowledge result, for instance, \textit{“Harris”} or \textit{“Trump”}, computed as \( v_* = \arg\min_v \mathcal{L}(v) \), where \( \mathcal{L}(v) \) is given by:

\begin{equation}
\label{eq7}
\mathcal{L}(v) = \frac{1}{N} \sum_{j=1}^N -\log P_{(v=v_*)}[o^* | p_j+x] + D_{KL}\left(P_{G(v=v_*)}[x | p'] \parallel P_G[x | p']\right).
\end{equation}

The first term serves to update the knowledge, while the second term preserves the essence of the subject. The objective is to modify the model's response to the knowledge query, yielding an output \( o^* \) (e.g., \textit{“Harris”} or \textit{“Trump”}). Additionally, \( C = KK^T \) is a pre-computed constant that estimates the uncentered covariance of \( k \), and \( \Lambda = (v_* - Wk_*)/(C^{-1} k_*)^T k_* \) represents the residual error of the new key-value pair. Further details can be found in \citep{meng2023locatingeditingfactualassociations}.

To manage editing intensity, \citep{meng2023masseditingmemorytransformer} introduced \textbf{MEMIT}, which computes matrix updates by solving:

\begin{equation}
\label{eq8}
\widetilde{W} = W + Rk_*^T(C + k_*k_*^T)^{-1},
\end{equation}

where \( \Delta W = Rk_*^T(C + k_*k_*^T)^{-1} \), \( C = \lambda \cdot KK^T \), and \( R = v_* - Wk_* \in \mathbb{R}^{d} \) is a column vector. The parameter \( \lambda \) allows for adjusting the balance between new edits and the original knowledge. It is noteworthy that in both ROME and MEMIT, \textbf{only \( v_* \) is derived through the training process}, and this operation will be optimized in subsequent steps. For additional implementation details regarding MEMIT, please refer to Appendix \ref{memit}.

\subsection{Towards an orthogonal editing method}
Previous methods share a common feature: all new knowledge is updated within a shared space, which directly affects the weights of the model. If an update for new knowledge is applied without considering prior knowledge, the direction of this update can affect both the previously edited knowledge and the implicit knowledge within the model, potentially leading to catastrophic forgetting \citep{luo2024empiricalstudycatastrophicforgetting, wang2023comprehensivesurveyforgettingdeep}. Therefore, to effectively support sequential editing, the process of updating new knowledge should adhere to the following criteria:

\textbf{Criterion 3.1:} \label{cer1} The update direction for each piece of knowledge should be orthogonal to the directions of previously edited knowledge, ensuring minimal interference with previously edited knowledge.

\textbf{Criterion 3.2:} \label{cer2} The update direction for each piece of knowledge should be orthogonal to the implicit knowledge directions within the original model, ensuring minimal interference with the model's existing implicit knowledge.

In the following sections \ref{3.2.1} and \ref{3.2.2}, we will detail how we optimized ROME and MEMIT to fulfill the two criteria mentioned above within the context of sequential editing.

\subsubsection{The knowledge to be edited should be mutually orthogonal}
\label{3.2.1}
\textbf{Editing the First Piece of Knowledge:} To comply with criterion 3.1, we implement the following steps in a sequential editing process. We commence by editing the first piece of knowledge using the pair \((x_1, y_1)\). Upon completion of this initial edit, we obtain an updated set of parameters \(\Delta W_{\text{[total]}} = \Delta W_{[1]}\). To preserve this edited knowledge, we constrain the gradient update directions for subsequent edits. It is important to note that during the editing process with methods such as ROME and MEMIT, parameter adjustments are made without gradient computation, as the calculation of \(v_*\) necessitates training, while the adjustment of $W_proj$ occurs in a single step. Since ROME and MEMIT do not involve computing the gradient direction of the required update matrix, we draw on the insights from \citep{wang2023orthogonalsubspacelearninglanguage} and utilize $\Delta W_\text{[total]}$ to approximate the direction of model parameter updates. They argue that the gradient space from prior training tasks can be effectively captured by the update matrix. Next, we perform Singular Value Decomposition (SVD) on \(\Delta W_{\text{[total]}} = U\Sigma V^T\) and extract the sub-matrix \(\Delta W_{r}\) corresponding to the top \(r\) singular values, defined as the Core Gradient Space (CGS) by \citep{saha2021gradientprojectionmemorycontinual}. Updates along the CGS direction induce maximum changes in knowledge \citep{farajtabar2019orthogonalgradientdescentcontinual}, whereas updates in directions orthogonal to the CGS minimize interference with previously edited knowledge\footnote{For additional details on updating within orthogonal subspaces, please refer to Appendix \ref{conlearning} and \ref{ograd}.}.

\textbf{Editing the Subsequent Knowledge:} To edit the second piece of knowledge using examples from \(D_{\text{edit}}\), we first retrieve the bases of the Core Gradient Space (CGS). The new update direction must lie in the space orthogonal to the CGS:

\vspace{-0.4cm}
\begin{equation}
\label{eqorth}
 \Delta W^T_{r} \cdot \Delta W_{[2]} = \mathbf{0}.
\end{equation}
\vspace{-0.4cm}

This ensures that the column vector subspace of \(W_2\) is orthogonal to the column vector subspace of \(W_r\). Taking MEMIT as an example, the update in Eq.\ref{eq8} can be optimized as\footnote{Since Eqn.\ref{eq5} involves \textbf{matrix right multiplication}, \(d\) denotes the column dimension and \(d_m\) denotes the row dimension.}:
\begin{equation}
\begin{aligned}
\label{eq10}
\widetilde{W} &= W + (v_* - Wk_*)k_*^T(C + k_*k_*^T)^{-1}, \\
\text{where} & \quad \Delta W^{T}_r \cdot (v_* - Wk_*)k_*^T(C + k_*k_*^T)^{-1} = \mathbf{0}.
\end{aligned}
\end{equation}
\vspace{-0.4cm}

Non-trivial solutions that approximately satisfy Eqn.\ref{eq10} can be obtained by training \(v_*\), where Eqn.\ref{eq7} can be rewritten as:

\begin{equation}
\begin{aligned}
\label{eq11}
\mathcal{L}(v) + \lambda_1 f_1(\Delta W_{r};v).
\end{aligned}
\end{equation}

Here:
\begin{equation}
\label{eq12}
\begin{aligned}
f_1 = \text{sim}\left(\Delta W_{r}, (v_* - Wk_*)k_*^T(C + k_*k_*^T)^{-1} \right),
\end{aligned}
\end{equation}

\textbf{\(\text{sim}\)} represents the cosine similarity function in column vector space, where each column vector lies in \(\mathbb{R}^{d}\), and \(\lambda_1\) serves as a hyperparameter that regulates the degree of orthogonality. Upon completion of the training of \(v_*\), Eqn. \ref{eq8} is employed to determine the update parameter \(\Delta W_{[2]}\). Following the update of the second piece of knowledge, the edited parameters are revised as follows:

\begin{equation}
\Delta W_{\text{[total]}} += \Delta W_{[2]}.
\end{equation}

We then proceed to the next piece of new knowledge, repeating the same procedure as for the second piece. The value of \(r\) increases linearly with each iteration of knowledge editing, defined as \(r = \text{min}(1 \times \text{Iteration}, \text{rank}(\Delta W_{\text{[total]}}))\). We provide an efficient solution for Eqn. \ref{eq12} and an explanation for \(r\) in Appendix \ref{o-edit}.

\subsubsection{The edited knowledge should be orthogonal to the implicit knowledge}
\label{3.2.2}
To adhere to criterion 3.2, we implement the following steps in the sequential editing process. We perform backpropagation on a large corpus of text to capture the model’s gradient information for the update direction of its internal implicit knowledge while freezing the original model's (unedited) parameters, simulating the pre-training process without updating the model, as illustrated in the bottom right of Figure \ref{fig2}. This computation is conducted on Wikipedia text, accumulating the gradient information by summing it. Appendix \ref{G} provides a comparison for selecting the appropriate text. Notably, this involves actual gradient information rather than the approximate update direction used in Section \ref{3.2.1}.

Once the gradient information \( \nabla G \in \mathbb{R}^{d \times d_m} \) of the implicit knowledge is obtained, the update direction for knowledge editing should be orthogonal to \( \nabla G \). Similar to Section \ref{3.2.1}, we obtain the rank \( q \) approximation of \( \nabla G \), denoted as \( \nabla G_q \), through SVD. We then subtract the projection of \( \nabla G_q \) onto \( W_r \) from \( \nabla G_q \):

\begin{equation}
\label{eq13}
\nabla G_q = \nabla G_q - \Delta W_r(\Delta W^T_r \Delta W_r)^{-1} \Delta W^T_r \nabla G_q,
\end{equation}

to prevent knowledge conflicts \citep{xu2024knowledgeconflictsllmssurvey, jin2024cuttingheadendsconflict} between the two. For instance, if \(\Delta W_r\) contains the edited knowledge \textit{"The President of the US is Harris/Trump"}, while \( \nabla G_q \) contains \textit{"The President of the US is Biden"}, the update directions for these two pieces of knowledge may conflict or even be completely opposite. In such cases, we prioritize preserving the knowledge in \(\Delta W_r\) over \( \nabla G_q \). The ultimate training objective is:

\begin{equation}
\begin{aligned}
\label{eq14}
\text{loss} = \mathcal{L}(z) + \lambda_1 f_1(\Delta W_r;v) + \lambda_2 f_2(\nabla G_q;v),
\end{aligned}
\end{equation}

where:

\vspace{-0.4cm}
\begin{equation}
\label{eq15}
\begin{aligned}
f_2 = \text{sim}\left(\nabla G_{q}, (v_* - Wk_*)k_*^T(C + k_*k_*^T)^{-1} \right).
\end{aligned}
\end{equation}
\vspace{-0.4cm}

The rank \(q\) increases linearly with the number of iterations of knowledge editing, described by \( q = \lambda_3 \times \text{iteration} \), where \( \lambda_3 \) is a hyperparameter controlling the degree of constraints.

Eqn. \ref{eq15} represents the final optimization target. After obtaining \( v_* \), we use Eqn. \ref{eq8} to solve for the update parameter. We then update the hyperparameters \( r \), \( q \), and \( \Delta W_{\text{[total]}} \) for the next knowledge update.

\section{\textbf{O-EDIT+}: Towards More Efficient Sequential Model Editing}
\label{4}
In Section \ref{3}, we introduced O-Edit, an algorithm for approximate orthogonal sequential knowledge editing. To further enhance the orthogonality between different pieces of knowledge, we propose O-Edit+, a post-processing method that eliminates the need for cosine similarity calculations. Specifically, for the second piece of knowledge, we compute \( v_* \) using Eqn.\ref{eq7} and apply Eqn.\ref{eq8} to obtain the update parameter \( \Delta W_{[2]} \). Subsequently, \( \Delta W_{[2]} \) undergoes post-orthogonal processing, achieved as follows:

\vspace{-0.4cm}
\begin{equation}
\label{eq16}
\begin{aligned}
\Delta W_{[2]} &= \Delta W_{[2]} - \Delta W_r (\Delta W^T_r \Delta W_r)^{-1} \Delta W^T_r \Delta W_{[2]},\\
\nabla G_q &= \nabla G_q - \Delta W_r(\Delta W^T_r \Delta W_r)^{-1} \Delta W^T_r \nabla G_q,\\
\Delta W_{[2]} &= \Delta W_{[2]} - \nabla G_q (\nabla G^T_q \nabla G_q)^{-1} \nabla G^T_q \Delta W_{[2]}.
\end{aligned}
\end{equation}

The processed \(\Delta W_{[2]}\) from Eqn.\ref{eq16} is then used as the update direction for the second piece of knowledge. Similar to O-Edit, we subsequently update the hyperparameters \( r \), \( q \), and \( \Delta W_{\text{[total]}} \) for the next knowledge edit. We detail the computation process of Eqn.\ref{eq16} and the pseudo-code for O-Edit and O-Edit+ in Appendix \ref{o-edit}. Readers can refer to Appendices \ref{oeditsettings} and \ref{Ablation} for details on hyperparameter selection.

\section{\textbf{Experiments}}

\begin{table}[t]
\vspace{-1cm}

    \caption{\textbf{Main editing results for COUNTERFACT.} $T$: Num Edits.}
    \centering
    % \setstretch{1.2}
    \resizebox{\linewidth}{!}{
    \begin{tabular}{lccc|c|ccc|c|ccc|c|ccc|c}
    \toprule
    \multirow{3}{*}{\textbf{Method}} & \multicolumn{16}{c}{\textbf{COUNTERFACT}} \\
     \cmidrule(lr){2-17}
     & \multicolumn{4}{c|}{$T=200$} & \multicolumn{4}{c|}{$T=500$} & \multicolumn{4}{c|}{$T=1000$} & \multicolumn{4}{c}{$T=1500$} \\
     \midrule
     & Rel. & Gen. & Loc. & Avg. & Rel. & Gen. & Loc. & Avg.  & Rel. & Gen. & Loc. & Avg. & Rel. & Gen. & Loc. & Avg. \\
     \midrule
     \multicolumn{17}{c}{\texttt{\textbf{Mistral-7B}}} \\
     \midrule
    \textbf{ROME} & 0.72 & 0.53 & 0.31 & 0.52 & 0.30 & 0.18 & 0.14 & 0.21 & 0.28 & 0.10 & 0.06 & 0.15 & 0.27 & 0.07 & 0.05 & 0.13 \\
    \cmidrule(lr){1-17}
     \qquad \textbf{+R-Edit} & 0.85 & \textbf{0.60} & 0.48 & 0.64 & 0.27 & 0.12 & 0.04 & 0.14 & 0.30 & 0.09 & 0.05 & 0.15 & 0.26 & 0.06 & 0.04 & 0.12 \\
     \qquad \textbf{+WilKE} & 0.81 & 0.59 & 0.44 & 0.61 & 0.45 & 0.27 & 0.19 & 0.30 & 0.28 & 0.10 & 0.10 & 0.16 & 0.18 & 0.02 & 0.07 & 0.09 \\
     \qquad \textbf{+PRUNE} & 0.76 & 0.51 & 0.28 & 0.52 & 0.35 & 0.21 & 0.21 & 0.26 & 0.42 & 0.12 & 0.05 & 0.20 & 0.33 & 0.15 & 0.22 & 0.23 \\
     \rowcolor{green!8}
     \qquad \textbf{+O-Edit} & \textbf{0.99} & 0.51 & 0.73 & \textbf{0.74} & \textbf{0.68} & 0.41 & 0.37 & 0.49 & 0.45 & 0.18 & 0.26 & 0.30 & 0.37 & 0.20 & 0.19 & 0.25 \\
     \rowcolor{red!8}
     \qquad \textbf{+O-Edit+} & 0.94 & 0.47 & \textbf{0.76} & 0.72 & 0.65 & \textbf{0.38} & \textbf{0.41} & \textbf{0.48} & \textbf{0.49} & \textbf{0.21} & \textbf{0.29} & \textbf{0.33} & \textbf{0.41} & \textbf{0.21} & \textbf{0.24} & \textbf{0.29} \\
     \cmidrule(lr){1-17}
     \textbf{MEMIT} & 0.93 & 0.67 & 0.41 & 0.67 & 0.50 & 0.35 & 0.10 & 0.32 & 0.28 & 0.10 & 0.06 & 0.15 & 0.19 & 0.06 & 0.05 & 0.10 \\
     \cmidrule(lr){1-17}
     \qquad \textbf{+R-Edit} & 0.93 & \textbf{0.64} & 0.48 & 0.68 & 0.76 & 0.39 & 0.16 & 0.44 & 0.32 & 0.17 & 0.06 & 0.18 & 0.28 & 0.13 & 0.06 & 0.16 \\
     \qquad \textbf{+WilKE} & \textbf{0.95} & 0.70 & 0.50 & 0.72 & 0.73 & 0.51 & 0.26 & 0.50 & 0.26 & 0.16 & 0.06 & 0.16 & 0.30 & 0.14 & 0.04 & 0.16 \\
     \qquad \textbf{+PRUNE} & 0.83 & 0.53 & 0.47 & 0.61 & 0.76 & 0.52 & 0.29 & 0.52 & 0.65 & 0.45 & 0.22 & 0.44 & 0.43 & 0.27 & 0.12 & 0.27 \\
      \rowcolor{green!8}
      \qquad \textbf{+O-Edit} & 0.93 & 0.55 & 0.65 & 0.71 & \textbf{0.86} & 0.53 & 0.45 & 0.61 & \textbf{0.72} & \textbf{0.47} & 0.34 & 0.51 & 0.51 & 0.33 & 0.18 & 0.34 \\
      \rowcolor{red!8}
     \qquad \textbf{+O-Edit+} & 0.89 & 0.61 &\textbf{ 0.78} & \textbf{0.76} & 0.81 & \textbf{0.55} & \textbf{0.60} & \textbf{0.65} & 0.68 & 0.39 & \textbf{0.55} & \textbf{0.54} & \textbf{0.61} & \textbf{0.42} & \textbf{0.53} & \textbf{0.52} \\
     \midrule
     \multicolumn{17}{c}{\texttt{\textbf{Llama3-8B}}} \\
     \midrule
     \textbf{ROME} & 0.75 & 0.48 & 0.14 & 0.46 & 0.69 & 0.45 & 0.05 & 0.40 & 0.75 & 0.46 & 0.02 & 0.41 & 0.47 & 0.28 & 0.02 & 0.31 \\
     \cmidrule(lr){1-17}
     \qquad \textbf{+R-Edit} & 0.70 & 0.38 & 0.27 & 0.45 & 0.65 & 0.41 & 0.06 & 0.37 & 0.54 & 0.34 & 0.03 & 0.30 & 0.50 & 0.31 & 0.02 & 0.28 \\
     \qquad \textbf{+WilKE} & 0.77 & 0.44 & 0.33 & 0.51 & 0.55 & 0.42 & 0.03 & 0.33 & 0.66 & 0.45 & 0.02 & 0.38 & 0.71 & 0.49 & 0.02 & 0.41 \\
     \qquad \textbf{+PRUNE} & \textbf{0.90} & 0.57 & 0.33 & 0.60 & 0.77 & 0.50 & 0.24 & 0.50 & 0.83 & 0.41 & 0.21 & 0.48 & 0.81 & 0.35 & \textbf{0.19} & 0.45 \\
     \rowcolor{green!8}
     \qquad \textbf{+O-Edit} & 0.88 & \textbf{0.63} & 0.35 & \textbf{0.62} & 0.77 & 0.47 & 0.22 & 0.49 & 0.84 & 0.47 & 0.13 & 0.48 & 0.83 & 0.31 & 0.09 & 0.41 \\
     \rowcolor{red!8}
     \qquad \textbf{+O-Edit+} & 0.86 & 0.61 &\textbf{ 0.37} & 0.61 & \textbf{0.81} & \textbf{0.52} & \textbf{0.24} & \textbf{0.52} & \textbf{0.86} & \textbf{0.49} & \textbf{0.19} & \textbf{0.51} & \textbf{0.87} & \textbf{0.50} & 0.13 & \textbf{0.50} \\
    \cmidrule(lr){1-17}
     \textbf{MEMIT} & 0.85 & 0.51 & 0.22 & 0.52 & 0.50 & 0.35 & 0.10 & 0.32 & 0.28 & 0.10 & 0.05 & 0.14 & 0.18 & 0.06 & 0.05 & 0.10 \\
     \cmidrule(lr){1-17}
     \qquad \textbf{+R-Edit} & 0.92 & 0.63 & 0.48 & 0.68 & 0.57 & 0.39 & 0.15 & 0.37 & 0.34 & 0.17 & 0.06 & 0.19 & 0.27 & 0.13 & 0.05 & 0.15 \\
     \qquad+ \textbf{WilKE} & \textbf{0.95} & \textbf{0.68} & 0.50 & 0.71 & 0.71 & \textbf{0.56} & 0.25 & 0.51 & 0.30 & 0.16 & 0.08 & 0.18 & 0.30 & 0.14 & 0.05 & 0.16 \\
     \qquad \textbf{+PRUNE} & 0.82 & 0.52 & 0.47 & 0.60 & 0.76 & 0.52 & 0.38 & 0.55 & 0.64 & 0.44 & \textbf{0.32} & 0.47 & 0.42 & 0.27 & 0.22 & 0.30 \\
     \rowcolor{green!8}
     \qquad \textbf{+O-Edit} & 0.93 & 0.55 & 0.64 & 0.71 & \textbf{0.86} & 0.53 & 0.44 & 0.61 & \textbf{0.72} & \textbf{0.47} & 0.33 & 0.51 & 0.55 & 0.40 & 0.27 & 0.41 \\
     \rowcolor{red!8}
     \qquad+\textbf{O-Edit+} & 0.88 & 0.53 & \textbf{0.76} & \textbf{0.72} & 0.84 & 0.51 & \textbf{0.45} & \textbf{0.60} & 0.81 & 0.50 & 0.31 & \textbf{0.54} & \textbf{0.79} & \textbf{0.44} & \textbf{0.28} & \textbf{0.50} \\
     \bottomrule
    \end{tabular}
    }
    \label{counterfact}
    \vspace{-0.4cm}
\end{table}

\subsection{Editing Experimental Settings and Evaluation Metrics}

\textbf{Datasets and Models.} We utilize autoregressive LLMs, specifically \textbf{Mistral-7B} \citep{jiang2023mistral7b} and \textbf{Llama3-8B}\footnote{\url{https://llama.meta.com/llama3}}, for evaluation, along with the datasets \textbf{ZsRE} \citep{decao2021editingfactualknowledgelanguage} and \textbf{COUNTERFACT} \citep{meng2023locatingeditingfactualassociations}. The ZsRE dataset is a question-answering resource comprising examples that include a sentence to be edited, its back-translated equivalent, and an unrelated sentence. The COUNTERFACT dataset consists of counterfactual examples derived from Wikipedia, where the probability of the new target is lower than that of the original answer. Examples from both datasets are provided in Appendix \ref{Editing Datasets}.

\textbf{Baseline.} We selected \textbf{ROME} \citep{meng2023locatingeditingfactualassociations} and \textbf{MEMIT} \citep{meng2023masseditingmemorytransformer} as baseline editors and compared our alternative methods, \textbf{O-Edit} and \textbf{O-Edit+}, with the following approaches: \textbf{R-Edit} \citep{gupta2024rebuildingromeresolving}, which facilitates sequential editing by solving for a more accurate \( k_* \); \textbf{WilKE} \citep{hu2024wilkewiselayerknowledgeeditor}, which dynamically selects the optimal layer for editing; and \textbf{PRUNE} \citep{ma2024perturbationrestrainedsequentialmodelediting}, which aims to enhance model robustness in sequential editing by adjusting the magnitude of singular values in \( \Delta W_{\text{[total]}} \), requiring prior edits to accumulate to a predetermined number before processing. Details on these methods are provided in Appendix \ref{baseline}.

\textbf{Metrics.} Each edit example comprises an edit knowledge statement, consisting of an edit statement \( \mathbf{x_e} \) and an edit target \( \mathbf{y_e} \), its paraphrase sentences \( \mathbf{x_{e'}} \) for testing generalization, and an unrelated knowledge statement \( \mathbf{x_{loc}} \) for testing locality. For the editing dataset \( \mathcal{D}_{\text{edit}} = \{(\mathbf{x_e}, \mathbf{y_e})\} \) with \( T \) edits, we evaluate the final post-edit model \( f_{\Theta_T} \) after the \( T \)-th edit example \( (\mathbf{x_T}, \mathbf{y_T}) \). We assess the reliability and generalization of the model editor using the metrics \textbf{Rel.} (Edit Success Rate \citep{zhang2024comprehensivestudyknowledgeediting}) and \textbf{Gen.} (Generalization Success Rate), while \textbf{Loc.} (Localization Success Rate) evaluates specificity, defined as the post-edit model's ability to maintain the output of the unrelated knowledge \( \mathbf{x_{loc}} \). We report these metrics and their mean scores, which are formally defined as:

\vspace{-0.4cm}
\begin{scriptsize} 
\begin{equation}
\text{Rel.} = \frac{1}{T} \sum_{t=1}^{T} \mathds{1}(f_{\Theta_T}(\mathbf{x}_e^t) = \mathbf{y}_e^t), 
\text{Gen.} = \frac{1}{T} \sum_{t=1}^{T} \mathds{1}(f_{\Theta_T}(\mathbf{x}_{e'}^t) = \mathbf{y}_e^t), 
\text{Loc.} = \frac{1}{T} \sum_{t=1}^{T} \mathds{1}(f_{\Theta_T}(\mathbf{x}_{\text{loc}}^t) = f_{\Theta_0}(\mathbf{x}_{\text{loc}}^t)),
\end{equation}
\end{scriptsize}
\vspace{-0.4cm}

where $\mathds{1}(\cdot)$ denotes the indicator function, indicating that we only consider the top-1 token during inference.

\textbf{Main Results.} The competitive performance of our methods is demonstrated in Tables \ref{counterfact} and \ref{zsrezsre}, which compare results against four baselines in the COUNTERFACT and ZsRE settings. Our observations are as follows: \ding{172} ROME and MEMIT exhibit increased forgetting of both previously edited and unrelated knowledge as the number of edits rises. \ding{173} Some methods, although effective for a limited number of edits (up to 500), struggle to maintain a balance between edit quality and retention of unrelated knowledge as the number of edits increases. \ding{174} O-Edit and O-Edit+ consistently outperform other methods in sequential editing tasks.

In the \textbf{COUNTERFACT} setting, with \( T = 200 \), models edited with MEMIT and ROME still perform effective edits. However, as the number of edits exceeds 500, their performance declines rapidly. After 1,500 edits on Mistral-7B, MEMIT's scores dropped to approximately 0.20 for \textbf{Rel.} and 0.05 for \textbf{Loc.}, indicating substantial forgetting of both edited and unrelated knowledge. Although improved methods like PRUNE and WilKE showed competitive performance at \( T = 200 \), they similarly failed to maintain a good balance across \textbf{Rel.}, \textbf{Gen.}, and \textbf{Loc.} at \( T = 1500 \).

At \( T = \{500, 1000, 1500\} \), O-Edit and O-Edit+ achieved the best results on both Mistral-7B and Llama3-8B. At \( T = 1500 \) with Mistral-7B, O-Edit+ improved by 0.16 and 0.42 in \textbf{Avg.} over ROME and MEMIT, respectively, and by 0.06 and 0.25 over PRUNE, our closest competitor. Overall, while performance across methods is similar for smaller numbers of edits, O-Edit+ significantly reduces forgetting as the number of edits increases, effectively preserving both edited and unrelated knowledge. Results for ZsRE are provided in Figure \ref{zsrezsre} in the Appendix.

\subsection{Downstream Tasks Evaluation}

\textbf{Datasets.} To investigate the side effects of sequential model editing on the downstream task abilities of LLMs, we adopted four representative tasks with corresponding datasets for assessment: \textbf{Commonsense Reasoning} using the \textbf{SIQA} \citep{sap2019socialiqacommonsensereasoningsocial}, \textbf{Content Analysis} on the \textbf{LAMBADA} \citep{paperno2016lambadadatasetwordprediction}, \textbf{Question Answering} with the \textbf{CommonsenseQA} \citep{talmor2019commonsenseqaquestionansweringchallenge}, and \textbf{MATH} on the \textbf{GSM8K} \citep{cobbe2021trainingverifierssolvemath}. Details of the prompts for each task are provided in Appendix \ref{downstream}.

\begin{figure}
\vspace{-0.5cm}
    \centering
    \includegraphics[width=1\textwidth]{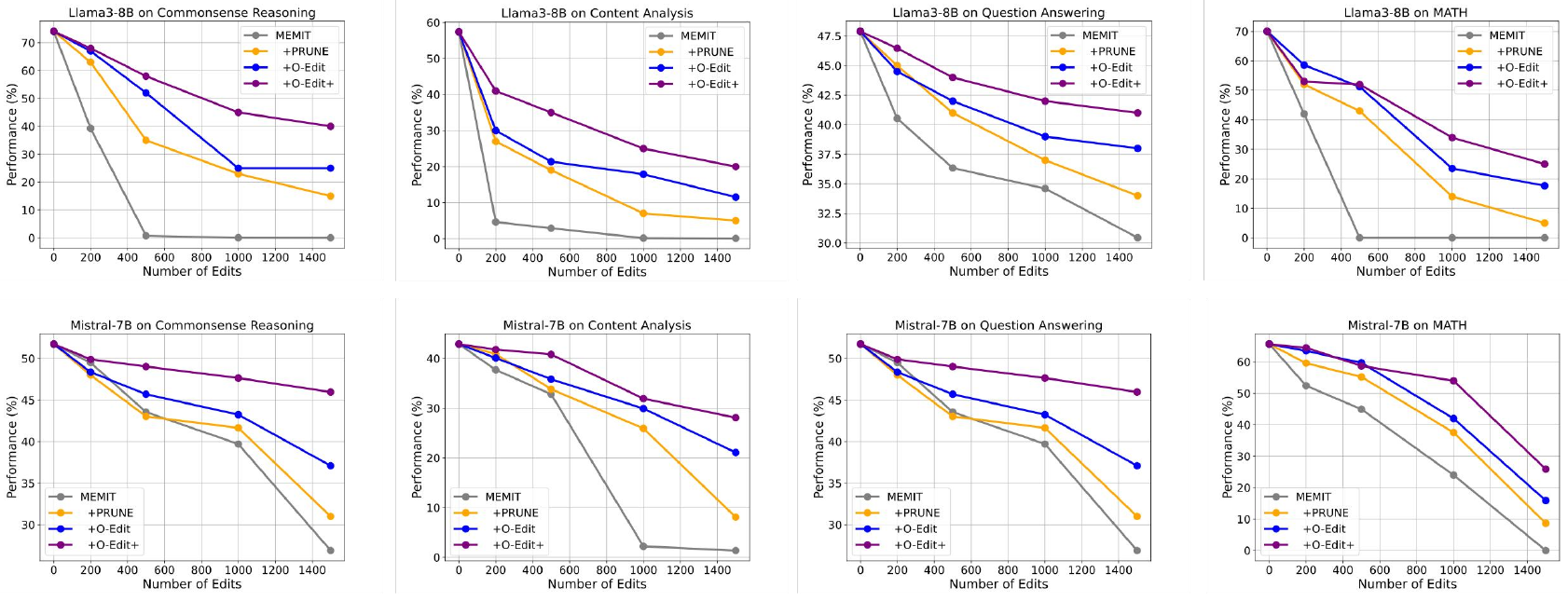}
    \caption{The downstream task performance (\%) of models edited by four editing methods with Mistral-7B and Llama3-8B on the COUNTERFACT dataset.}
    \label{fig3}
    \vspace{-0.5cm}
\end{figure}

\textbf{Main Results.} Figure \ref{fig3} illustrates the downstream task performance of Mistral-7B and Llama3-8B after applying MEMIT and O-Edit+ in the \textbf{COUNTERFACT} setting. As shown by the gray line in Figure \ref{fig3}, MEMIT maintains performance at a certain level when the number of edits is small ($T \leq 200$). However, as the number of edits exceeds 1000, MEMIT's performance drastically declines, approaching zero (with results on CommonsenseQA resembling random guessing, both around 20\%). In contrast, O-Edit and O-Edit+ effectively tackle this issue by implementing constraints that ensure orthogonality between the editing knowledge and the original model’s implicit knowledge, significantly reducing interference. With O-Edit+ applied for 200 edits, downstream task performance remains close to that of the unedited model, effectively preserving accuracy across various tasks. Even after 1,500 edits, O-Edit+ remains to outperform both MEMIT and PRUNE, demonstrating its robustness in maintaining downstream task performance over extended sequences of edits. This highlights the effectiveness of O-Edit+ in minimizing interference between edits, allowing models to retain high performance even in heavily edited environments.

Nevertheless, as the number of edits increases, extensive knowledge editing inevitably leads to diminished model performance, a phenomenon described by \citep{wang2024wiserethinkingknowledgememory} as the ``unbreakable triangle,'' which asserts that no method can achieve perfect editing without compromising other aspects of the model’s performance. Despite this, O-Edit+ significantly mitigates this effect, offering superior performance retention compared to other editing methods such as MEMIT.

\begin{figure}
    \vspace{-1cm}
    \centering
    \includegraphics[width=0.6\textwidth]{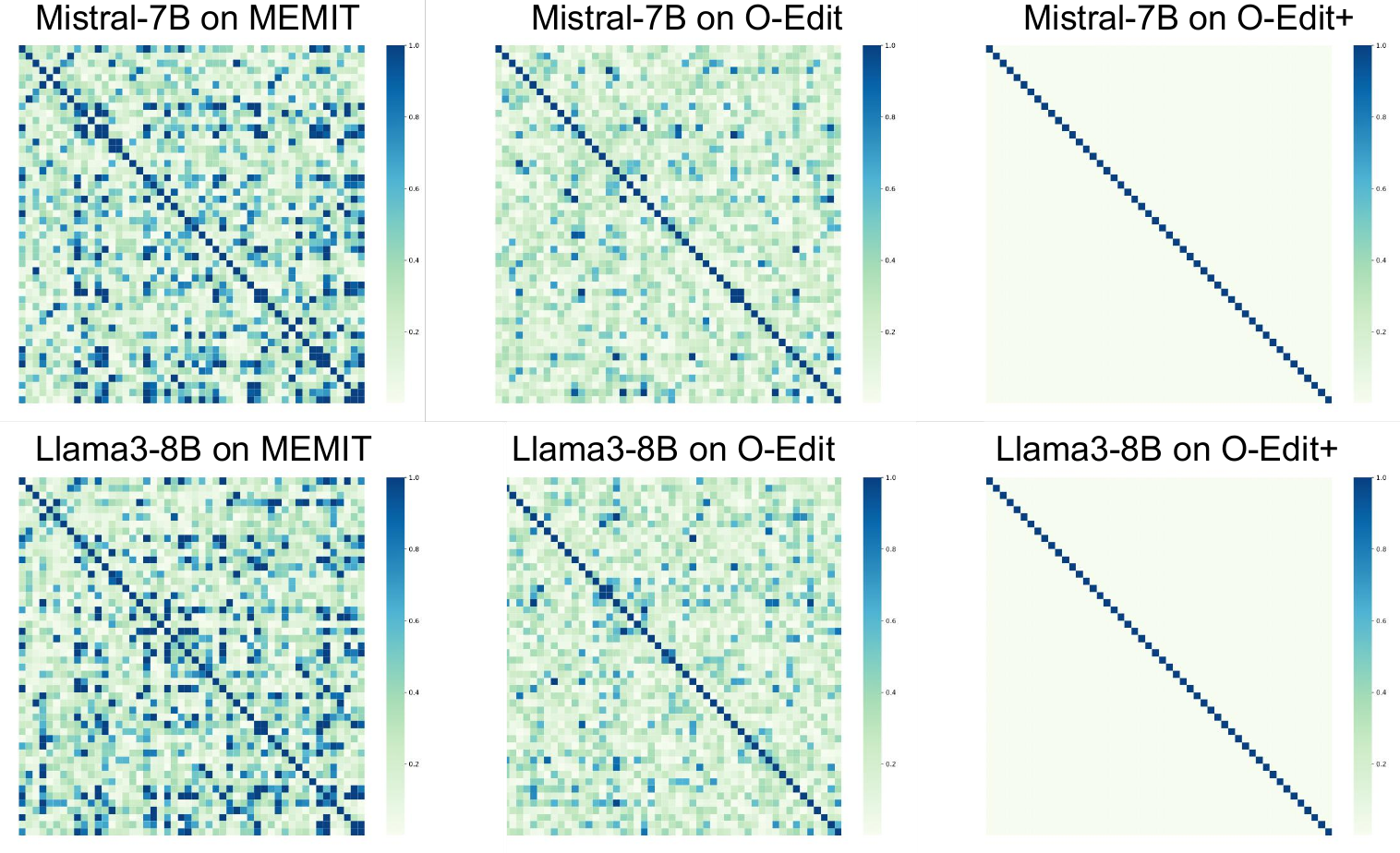}
        \caption{This figure illustrates the update directions of three editing methods on the COUNTERFACT dataset, with orthogonality values scaled by a factor of 10 for clarity. The horizontal and vertical axes represent the selected editing samples.}
        \label{fig4}
        \vspace{-0.5cm}
\end{figure}

\subsection{Further Analysis}
\label{Further Analysis}

\textbf{How does O-Edit alter the direction of updates for each knowledge component?} We evaluated the orthogonality among each update matrix, \(\Delta W\), by examining the cosine similarity between the corresponding matrices after applying MEMIT, O-Edit, and O-Edit+. To conduct this evaluation, we randomly selected 50 edits and computed the absolute cosine similarity between their corresponding \(\Delta W\) matrices. This was achieved by calculating the cosine similarity between the \(V\) matrices derived from each \(\Delta W\) after performing singular value decomposition. As illustrated in Figure \ref{fig4}, without any constraints, there is a significant overlap in the update directions, which may cause subsequent edits to influence the directions of prior edits. O-Edit mitigates this overlap by training an appropriate \(v_*\), while O-Edit+ achieves complete orthogonality between each update direction through post-processing.

\textbf{How do edits disturb each other?} To investigate the extent of interdependencies among knowledge updates during the sequential editing process, we preserved the update matrix \(\Delta W_i\) for each \(i\)-th edit. Upon completion of the sequential editing, the model's cumulative update matrix is computed as \(\Delta W_{\text{[total]}} = \sum_{i=1}^n \Delta W_i\). For the \(j\)-th edit, we define \(\Delta W_\text{unrelated} = \Delta W_{\text{[total]}} - \Delta W_j\), which excludes the update matrix \(\Delta W_j\) corresponding to \(k_j\). According to \cite{hu2024knowledgesuperpositionunveilingfailures}, under ideal sequential editing, the knowledge vector \(k_j\) used during the \(j\)-th edit should not activate any unrelated \(\Delta W_{\not = j}\) (i.e., any update matrix other than \(\Delta W_j\)), meaning \( \| \Delta W_\text{unrelated} \cdot k_j\|_2 = 0\). We calculate the activation score (AS) for each edit as \(\| \Delta W_\text{unrelated} \cdot k_j\|_2\).

\begin{wrapfigure}{r}{0.3\textwidth} 
  \vspace{-0.5cm}
  \centering
  \includegraphics[width=0.3\textwidth]{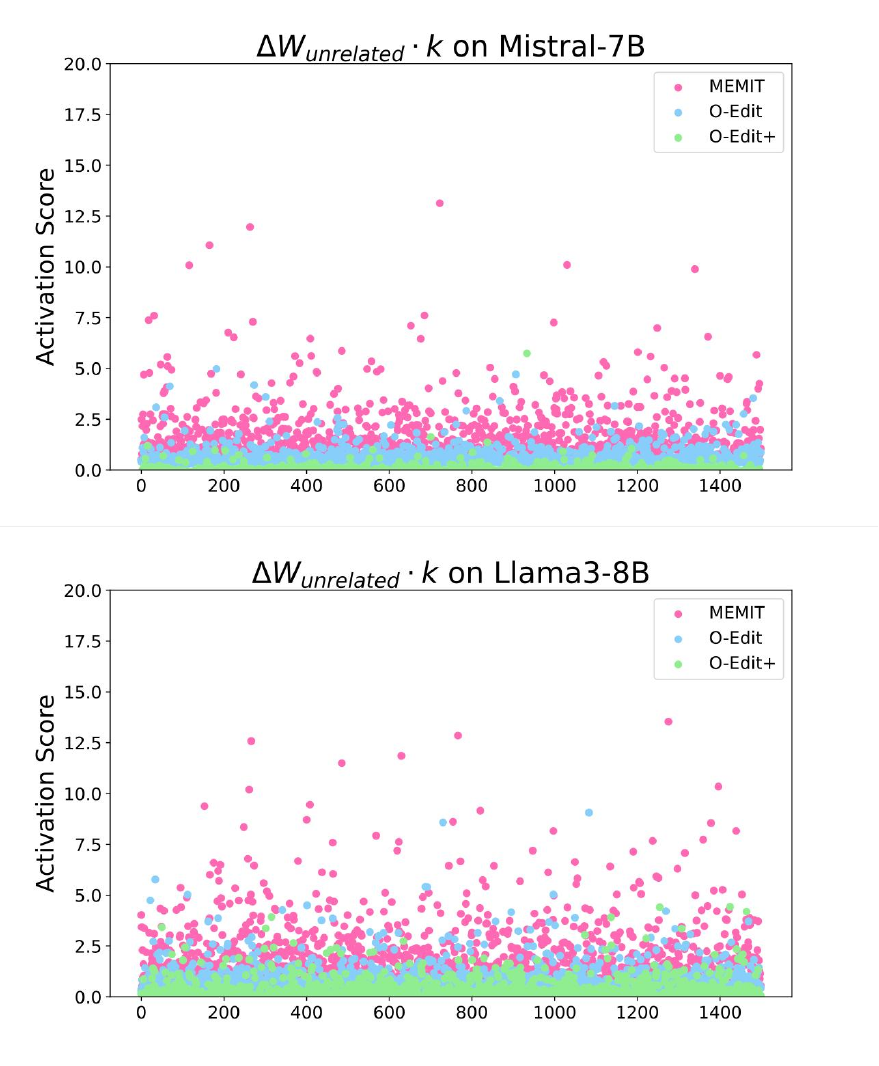}
  \caption{The activation score caused by unrelated parameters. \texttt{X-axis}: Number of edits.}
  \label{fig5}
  \vspace{-0.5cm}
\end{wrapfigure}

As illustrated in Figure \ref{fig5}, after 1,500 edits, the original method exhibited high activation scores (AS), with some values reaching approximately 2.5 and others exceeding 10. This indicates that in the original method, any unrelated \(\Delta W_{\not = j}\) (i.e., any update matrix other than \(\Delta W_j\)) could significantly activate \(k_j\), leading to a substantial deviation from the ideal state \(v_*\) and resulting in the failure of MEMIT in sequential editing. In contrast, both O-Edit and O-Edit+ consistently achieved activation values below 2.5 for nearly all edits, with some values approaching zero. In Appendix \ref{different}, we analyze the reasons for this phenomenon from a mathematical derivation perspective, highlighting that the key lies in the orthogonality of the column subspaces of each update matrix.

\textbf{Can any method of reducing \(\Delta W_{\text{[total]}}\) improve the ability of sequential editing?} \cite{hu2024wilkewiselayerknowledgeeditor} posits that \(\|\Delta W_{\text{[total]}}\|_2\) is a key determinant of sequential editing, referred to as \textbf{``toxicity''}. A higher \(\|\Delta W_{\text{[total]}}\|_2\) imposes greater constraints on sequential editing performance. O-Edit+ effectively reduces \(\|\Delta W_{\text{[total]}}\|_2\) by diminishing projections in specific subspaces. Consequently, a plausible hypothesis is that any method capable of reducing \(\|\Delta W_{\text{[total]}}\|_2\) could potentially enhance sequential editing performance. To evaluate this hypothesis, we compare O-Edit+ with four methods on COUNTERFACT: \ding{202} reducing the number of training steps to decrease \(\|v_* - Wk_*\|_2\), thereby reducing \(\|\Delta W_{\text{[total]}}\|_2\) with each edit; \ding{203} randomly deleting some values in the update parameters, setting them to zero; \ding{204} randomly selecting a set of orthogonal subspaces and removing the projection of \(\Delta W_i\) onto them; \ding{205} multiplying the \(\Delta W\) obtained by the original method by a coefficient \(\eta\) that is less than 1, updating the matrix as \(\Delta W =\eta \cdot \Delta W \). We adjust the hyperparameters to ensure that the \(\|\Delta W_{\text{[total]}}\|_2\) generated by these methods approximates that of O-Edit+. As shown in Table \ref{table2} and Appendix Table \ref{canany}, although these five methods yield a similar \(\|\Delta W_{\text{[total]}}\|_2\), the first four fail to achieve effective sequential editing. This indicates that while reducing \(\|\Delta W_{\text{[total]}} \|_2\) is a necessary but not sufficient condition for successful sequential editing, choosing the correct projection space to ensure minimal impact between knowledge is the key to the success of ours.

\begin{table}[ht]
    \begin{minipage}[t]{0.45\textwidth} % 左侧表格
        \centering
        \scriptsize % 设置表格字体
        \caption{Comparison of different methods under two scenarios, $T$ = \{500, 1500\}} 
        \label{table2}
        \begin{tabular}{lccccc}
        
        \toprule
        \multirow{2}{*}{Mistral-7B} & \multicolumn{5}{c}{$T=500$} \\
        \cmidrule(lr){2-6}
         & Rel. & Gen. & Loc. & Avg. &\(\|\cdot \|_2\) \\
        \midrule
        MEMIT & 0.50 & 0.35 & 0.10 & 0.32 &\textbf{168}\\
        Method \ding{202} & 0.41 & 0.22 & 0.44 & 0.36 &106 \\
        Method \ding{203} & 0.57 & 0.34 & 0.31 & 0.40 & 102\\
        Method \ding{204} & 0.60 & 0.37 & 0.30 & 0.42 &109\\
        Method \ding{205} & 0.57 & 0.33 & 0.31 & 0.40 &102\\
        O-Edit+ & \textbf{0.81} & \textbf{0.55} & \textbf{0.60} & \textbf{0.65} &109 \\
        \midrule
        \multirow{2}{*}{Mistral-7B} & \multicolumn{5}{c}{$T=1500$} \\
        \cmidrule(lr){2-6}
         & Rel. & Gen. & Loc. & Avg. &\(\|\cdot \|_2\)\\
        \midrule
        MEMIT & 0.19 & 0.06 & 0.05 & 0.10  &\textbf{681}\\
        Method \ding{202} & 0.20 & 0.08 & 0.09 & 0.12 &355\\
        Method \ding{203} & 0.22 & 0.13 & 0.04 & 0.13 &367 \\
        Method \ding{204} & 0.18 & 0.10 & 0.06 & 0.11 &366\\
        Method \ding{205} & 0.21 & 0.11 & 0.05 & 0.12 &359\\
        O-Edit+ & \textbf{0.61} & \textbf{0.42} & \textbf{0.53} & \textbf{0.52} &361\\
        \bottomrule
        \end{tabular}
    \end{minipage}
    \hfill % 插入空白分隔两张表格
    \begin{minipage}[t]{0.45\textwidth} % 右侧表格

        \centering
        \scriptsize % 设置表格字体
        \caption{Comparison of different hyperparameters in O-Edit.} 
        \label{oedithy}
        \begin{tabular}{lcccc}
        \toprule
        \multirow{2}{*}{Mistral-7B} & \multicolumn{4}{c}{$T=1500$} \\
        \cmidrule(lr){2-5}
         & Rel. & Gen. & Loc. & Avg.  \\
        \midrule
         \(\lambda_1\), \(\lambda_2\)=0 & 0.19 & 0.06 & 0.05 & 0.10 \\
         \(\lambda_1\), \(\lambda_2\)=1 & 0.39 & 0.19 & 0.08 & 0.22 \\
        \(\lambda_1\), \(\lambda_2\)=10 & 0.45 & 0.22 & 0.10 & 0.26 \\
        \(\lambda_1\), \(\lambda_2\)=20 & 0.47 & 0.26 & 0.13 & 0.29 \\
        \(\lambda_1\), \(\lambda_2\)=50 & \textbf{0.51} & \textbf{0.33} & \textbf{0.18} & \textbf{0.34}  \\
        \midrule
        \multirow{2}{*}{Llama3-8B} & \multicolumn{4}{c}{$T=1500$} \\
        \cmidrule(lr){2-5}
         & Rel. & Gen. & Loc. & Avg. \\
        \midrule
         \(\lambda_1\), \(\lambda_2\)=0 & 0.18 & 0.06 & 0.05 & 0.10 \\
         \(\lambda_1\), \(\lambda_2\)=1 & 0.35 & 0.27 & 0.08 & 0.23 \\
        \(\lambda_1\), \(\lambda_2\)=10 & 0.45 & 0.33 & 0.10 & 0.29 \\
        \(\lambda_1\), \(\lambda_2\)=20 & 0.45 & 0.35 & 0.15 & 0.31 \\
        \(\lambda_1\), \(\lambda_2\)=50 & \textbf{0.55} & \textbf{0.40} & \textbf{0.27} & \textbf{0.41}  \\
        \bottomrule
        \end{tabular}
    \end{minipage}
    \vspace{-0.3cm}
\end{table}

\textbf{How does the degree of orthogonality between knowledge affect the effectiveness of sequential editing?} To explore the impact of orthogonality on editing effectiveness, we modified the hyperparameters \(\lambda_1\) and \(\lambda_2 \in \{0, 1, 10, 20, 50\}\) in O-Edit to adjust the degree of orthogonality between knowledge. As shown in Table \ref{oedithy}, increasing \(\lambda_1\) and \(\lambda_2\) improves the orthogonality between knowledge (see Eqn.\ref{eq14}), leading to enhanced editing performance. When \(\lambda_1\) and \(\lambda_2\) are set to 0, O-Edit degenerates into MEMIT. Even with \(\lambda_1 = 1\), O-Edit outperforms MEMIT across all metrics. This improvement occurs because increased orthogonality ensures that edits are performed in independent subspaces, minimizing interference with previously edited knowledge and reducing catastrophic forgetting. When \(\lambda_1\) and \(\lambda_2\) reach 50, performance peaks, demonstrating an improvement of 0.24 in Mistral-7B and 0.31 in Llama3-8B compared to MEMIT. Higher values of \(\lambda_1\) and \(\lambda_2\) enhance the independence between editing directions, ensuring that new knowledge edits do not disrupt existing model knowledge. This indicates a positive correlation between orthogonality and editing effectiveness. For a more detailed comparison of hyperparameters in O-Edit and O-Edit+, please refer to Appendix \ref{Ablation}.

\vspace{-5pt}
\section{Limitations}
\vspace{-5pt}
While O-Edit and O-Edit+ demonstrate robust sequential editing performance, several limitations persist. Due to computational constraints, we restricted our experiments to Mistral-7B and Llama3-8B, leaving the scalability of our methods on larger models untested. Additionally, constructing orthogonality between edits adds computational overhead, which may prolong editing times. However, O-Edit and O-Edit+ require maintaining only two additional matrices, making them both model-agnostic and compatible with other sequential editing techniques. Furthermore, we did not evaluate O-Edit and O-Edit+ against other editing methods, such as fine-tuning (FT), as these approaches tend to falter after only a few sequential edits, whereas ROME and MEMIT can support more extensive editing sequences. Despite these challenges, we believe our methods hold promising potential, particularly in the early stages of research on sequential model editing.

\vspace{-5pt}
\section{Conclusion}
\vspace{-5pt}
In this paper, we present two innovative methods—O-Edit and O-Edit+ that leverage orthogonal subspace editing for sequential knowledge editing in language models. These methods effectively mitigate catastrophic forgetting of both edited and existing knowledge by incrementally applying edits in orthogonal subspaces. Our methods distinguish themselves through their attention to data privacy, efficient parameter utilization, and strong generalization capabilities for downstream tasks. Comprehensive empirical evaluations indicate that O-Edit and O-Edit+ significantly outperform existing methods, establishing them as promising avenues for future advancements in sequential knowledge editing.

\bibliography{iclr2025_conference}
\bibliographystyle{iclr2025_conference}

\appendix

\newpage
\section{Related Work}
\label{related work}
\subsection{ Knowledge Editing}
From the perspective of whether model parameters are modified, \citep{yao2023editinglargelanguagemodels} categorized knowledge editing methods into two major classes: preserving the model's parameters and modifying the model's parameters. This paper primarily focuses on the latter. On one hand, meta-learning has been used to predict parameter updates for networks, typically employing a hypernetwork to edit language models. \citep{decao2021editingfactualknowledgelanguage} used a bidirectional LSTM to predict weight updates for editing, while \citep{mitchell2022fastmodeleditingscale} utilized low-rank decomposition of gradients to fine-tune language models, known as MEND, and \citep{tan2024massiveeditinglargelanguage} extended single-step edits to batch edits using a least squares method based on MEND. On the other hand, \citep{meng2023locatingeditingfactualassociations, dai2022knowledgeneuronspretrainedtransformers} employed a causal probe to localize knowledge within the intermediate layers of the model, demonstrating that editing in the MLP of the middle layers yields the best results. \citep{dai2022knowledgeneuronspretrainedtransformers} performed knowledge editing by modifying the activation values of specific neurons. \citep{meng2023locatingeditingfactualassociations} used a constrained least squares method to precisely solve for the parameter updates required for editing and extended this approach to batch editing \citep{meng2023masseditingmemorytransformer}. 

\subsection{Sequential Editing}
Some studies have extended knowledge editing methods to sequential editing. From the perspective of modifying model parameters, \citep{ma2024perturbationrestrainedsequentialmodelediting} theoretically analyzed that the bottleneck limiting sequential editing in models lies in the condition number of matrices, and they attempted to support sequential editing by controlling the growth of the matrix condition number. \citep{hu2024wilkewiselayerknowledgeeditor} attributed the decline in performance during sequential editing to pattern mismatch, where different layers detect different patterns, making a single layer incapable of accommodating all the edited knowledge. Thus, they selected the optimal layer from multiple layers for editing. Additionally, \citep{hu2024knowledgesuperpositionunveilingfailures} explored the root causes of failures in sequential editing, deriving a closed-form solution from linear associative memory. They posited that lossless sequential editing can only be achieved when the edited knowledge is completely orthogonal. From the perspective of adding additional parameters while freezing model parameters, SERAC \citep{mitchell2022memorybasedmodeleditingscale} stores edits in memory. When an input is received, a classifier checks whether it corresponds to any cached edits. If a match is found, a counterfactual model uses the input and relevant edits to predict outputs. GRACE \citep{hartvigsen2023aginggracelifelongmodel} uses semantic similarity in the model's latent space by adding an offline key-value adapter at the selected layers, applying edits only to inputs that are similar to the keys cached in the encoding. WISE \citep{wang2024wiserethinkingknowledgememory} uses a dual-parameter storage scheme, where the main memory is used for pre-trained knowledge and the side memory is designated for edited knowledge. By incorporating a knowledge sharding mechanism, it allows for editing knowledge in different parameter subspaces and merges them into the shared side memory without causing conflicts. In this paper, we consider the scenario of directly updating model parameters.

\subsection{Continual learning}
\label{conlearning}
The orthogonal concept presented in this paper is inspired by continual learning. Existing continual learning methods typically update all tasks within a shared vector space \citep{ke2023continuallearningnaturallanguage}, which directly affects the model's hidden layer outputs \citep{wang2024comprehensivesurveycontinuallearning}. Some studies \citep{farajtabar2019orthogonalgradientdescentcontinual, saha2021gradientprojectionmemorycontinual} have proposed a promising approach to address this issue by performing gradient descent optimization in directions orthogonal to the gradient subspaces of past tasks, effectively mitigating catastrophic forgetting. GPM \citep{saha2021gradientprojectionmemorycontinual} divides the gradient space into two key areas: the ``Core Gradient Space'' (CGS) and the ``Residual Gradient Space'' (RGS). By learning in the orthogonal directions of the CGS related to previous task inputs, it ensures minimal interference with past tasks. Based on GPM, TRGP \citep{lin2022trgptrustregiongradient} introduces a ``trust region'' concept to select old tasks relevant to new ones, reusing their frozen weights through scaled weight projections. By optimizing the scaling matrix and updating the model along orthogonal directions to the old tasks' subspace, TRGP effectively facilitates knowledge transfer without forgetting. O-LoRA \citep{wang2023orthogonalsubspacelearninglanguage} suggests that parameter information updated through low rank can be approximately equivalent to gradient information, which expands the application scenarios of continual learning and enables effective learning even in scenarios where gradient information cannot be obtained.

\section{algorithm}

\subsection{Orthogonal Gradient Descent for Continual Learning}
\label{ograd}

Consider a continual learning setting where tasks $\{T_1, T_2, T_3, \ldots\}$ are learned sequentially without access to previous task data. Suppose that the model has been trained on \(T_A\) in the usual way until convergence to a update parameter \(w^*_A\). To mitigate the impact on $T_A$ while training on the next task $T_B$, \cite{farajtabar2019orthogonalgradientdescentcontinual} propose to ``orthogonalize'' it in a way that the new update direction \(\tilde{g}\) on \(T_B\) satisfies:

\begin{equation}
\tilde{g} \perp \nabla f(x; w^*_A), \quad \forall x \in T_A.
\end{equation}

One can compute and store \(\nabla f(x; w)\) for all \(x \in T_A\) when training on \(T_A\) is done. In a continual learning scenario involving multiple tasks, the direction of gradient updates is determined by:

\begin{equation}
    \tilde{g} = g - \sum_{i=1}^{n_A} \operatorname{proj}_{\mathbf{g}_i}(g) = g - \sum_{i=1}^{n_A} \left\langle g, \mathbf{g}_i \right\rangle \mathbf{g}_i
\end{equation}

The new direction -\(\tilde{g}\) is still a descent direction for \(T_B\), meaning that there exists \(\epsilon > 0\) such that for any learning rate \(0 < \eta < \epsilon\), taking the step -\(\eta \tilde{g}\) reduces the loss.

\subsection{Singular Value Decomposition and Rank-\texorpdfstring{$r$}{r} Approximation}
 Singular Value Decomposition (SVD) decomposes any matrix \( W \in \mathbb{R}^{m \times n} \) into three matrices: \( W = U \Sigma V^T \), where \( U \in \mathbb{R}^{m \times m} \) and \( V \in \mathbb{R}^{n \times n} \) are orthogonal matrices, and \( \Sigma \) is a diagonal matrix containing the singular values \( \sigma_i \) of \( W \), ordered in descending magnitude. SVD is instrumental in solving the rank-\( r \) approximation problem, where the goal is to find a matrix \( \tilde{W} \) that minimizes \( \|\tilde{W} - W\|_2 \) subject to \( \operatorname{rank}(\tilde{W}) \leq r \). According to the Eckart–Young–Mirsky theorem \citep{Eckart1936TheAO}, the optimal rank-\( r \) approximation \( \tilde{W} \) is given by \( \tilde{W} = \sum_{i=1}^r \sigma_i u_i v_i^T \), obtained by truncating the SVD of \( W \) to retain the top \( r \) singular values and their corresponding singular vectors, where \( r \leq \min\{m, n\} \).

\subsection{MEMIT}
\label{memit}

In this paper, we consider the scenario of editing one piece of knowledge at a time. Similar to ROME, MEMIT views \(W^{l}_{\text{proj}}\) as a linear key-value memory for a set of vector keys \(K = \{k_1, k_2, \ldots\}\) and corresponding vector values \(V = \{v_1, v_2, \ldots\}\) by solving \(WK = V\). It attempts to insert a new key-value pair \((k_*, v_*)\) into the model by solving the following constrained least squares problem:
\begin{equation}
\label{eq18}
\widetilde{W} = \underset{\hat{W}}{\operatorname{argmin}} \left( \left\|\hat{W} K - V\right\|_2 + \left\|\hat{W} k_* - v_*\right\|_2 \right).
\end{equation}

MEMIT solves Eqn. \ref{eq18} by applying the normal equation, which is expressed in block form:
\begin{equation}
\widetilde{W} \begin{bmatrix} K & k_* \end{bmatrix} \begin{bmatrix} K^T \\ k_*^T \end{bmatrix} = \begin{bmatrix} V & v_* \end{bmatrix} \begin{bmatrix} K^T \\ k_*^T \end{bmatrix}, \tag{10}
\end{equation}
which expands to:
\begin{equation}
(W + \Delta) \left( K K^T + k_* k_*^T \right) = V K^T + v_* k_*^T, \tag{11}
\end{equation}
\begin{equation}
W K K^T + W k_* k_*^T + \Delta K^T K + \Delta k_*^T k_* = V K^T + v_* k_*^T. \tag{12}
\end{equation}

Under the condition \(WK = V\), we can simplify to:
\begin{equation}
\Delta (K K^T + k_* k_*^T) = v_* k_*^T - W k_* k_*^T,
\end{equation}
yielding:
\begin{equation}
\Delta = (v_* - Wk_*) k_*^T (K K^T + k_* k_*^T)^{-1}.
\end{equation}

Thus, the final update rule is:
\begin{equation}
\widetilde{W} = W + (v_* - Wk_*) k_*^T (K K^T + k_* k_*^T)^{-1}.
\end{equation}

Here, \((v_* - Wk_*) \in \mathbb{R}^{d}\) is a column vector, and \(k_*^T (K K^T + k_* k_*^T)^{-1} \in \mathbb{R}^{d_m}\) is a row vector. By adjusting the hyperparameter \(\lambda\), MEMIT balances the preservation of existing knowledge and the incorporation of new edits. Consequently, the updated equation is expressed as follows:
\begin{equation}
\widetilde{W} = W + (v_* - Wk_*) k_*^T (\lambda K K^T + k_* k_*^T)^{-1}.
\end{equation}

Like ROME, \(KK^T\) is pre-cached by estimating the uncentered covariance of \(k\) from a sample of Wikipedia text. The rank of the update matrix \(\Delta W = (v_* - Wk_*) k_*^T (\lambda K K^T + k_* k_*^T)^{-1}\) obtained through ROME and MEMIT is 1.

In fact, MEMIT is a scalable extension of ROME. By increasing \(\lambda\), MEMIT effectively enhances the retention of existing knowledge while also allowing for new updates. However, the restrictive conditions imposed by ROME, which require \(k_* \widetilde{W} \equiv v_*\) as seen in Eqn. \ref{eq5}, can be overly stringent and may lead to greater disruption of existing knowledge within the model.

\subsection{O-Edit and O -Edit+}
\label{o-edit}
\textbf{We will provide a detailed explanation of the calculation formula for O-Edit.} To explain how to compute Eqn. \ref{eq12}, we first analyze the properties of the update matrices for each piece of knowledge. Based on the matrix property \(\text{rank}(AB) \leq \min(\text{rank}(A), \text{rank}(B))\), the ranks of \(\Lambda(C^{-1} k_*)^T\) in Eqn. \ref{eq5} and \(Rk_*^T(C + k_*k_*^T)^{-1}\) in Eqn. \ref{eq8} are both 1. In the \(i\)-th edit, the rank of the cached \(\Delta W_{\text{[total]}} \in \mathbb{R}^{d \times d_m}\) is at most \(i\), with equality when each \(k_*\) is linearly independent. After several edits, \(\text{rank}(\Delta W_{\text{[total]}}) = 1 \times \text{iteration}\), but as updates increase, \(\text{rank}(\Delta W_{\text{[total]}})\) may fall below the iteration count. Therefore, \(r\) is always equal to \(\text{rank}(\Delta W_{\text{[total]}})\), and \(\Delta W_r\) is \(\Delta W_{\text{[total]}}\) itself.

During the computation process, we observe that \(\Delta W^T_r = U_{\Delta W_r} \Sigma_{\Delta W_r} V^T_{\Delta W_r}\), where \(U_{\Delta W_r} \in \mathbb{R}^{d_m \times r}\), \(V_{\Delta W_r} \in \mathbb{R}^{d \times r}\), and \(\Sigma_{\Delta W_r}\) is a diagonal matrix. Eqn. \ref{eqorth} can be rewritten as:
\begin{equation}
\label{svdorth}
U_{\Delta W_r} \Sigma_{\Delta W_r} V^T_{\Delta W_r} \cdot \Delta W_{[2]} = \mathbf{0}.
\end{equation}
We only need to ensure that \(v_* - Wk_*\) is orthogonal to \(V_r\). Therefore, Eqn. \ref{eq12} can be rewritten as:
\begin{equation}
\begin{aligned}
\label{eqf1}
f_1 = \frac{1}{r} \sum_{i=0}^{r} \text{sim}(V_{\Delta W_r}[i], (v_* - Wk_*)).
\end{aligned}
\end{equation}

The key reason for using cosine similarity instead of \(V^T_{\Delta W_r} \cdot (v_* - Wk_*)\) is that the latter may lead to trivial solutions, i.e., \(v_* - Wk_* = \mathbf{0}\), while cosine similarity considers angular information. In fact, merely reducing the norm of \(v_* - Wk_*\) does not effectively enhance the effectiveness of sequential editing. The success of O-Edit and O-Edit+ lies in identifying the correct update direction during the sequential editing process. For further details, see \textbf{Further Analysis} \ref{Further Analysis}.

Furthermore, when calculating \(\nabla G\), we utilized a large amount of natural text, resulting in \(\nabla G\) being a high-rank matrix, which is distinct from \(\Delta W_{\text{[total]}}\). We dynamically adjust \(q\) to select the core gradient subspace (CGS) of \(\nabla G\), defined as \(\nabla G^T_q = U_{\nabla G_q} \Sigma_{\nabla G_q} V^T_{\nabla G_q}\). The purpose of this adjustment is to counteract the cumulative impact of edited knowledge on the implicit knowledge within the model as the number of edits increases. We adjust \(q\) to increase linearly with the number of edits. In practice, we compute Eqn. \ref{eq13} by removing the projection of \(V_{\nabla G_q}\) onto \(V_{\Delta W_r}\):
\begin{equation}
\label{ortheacho}
V_{\nabla G_q} = V_{\nabla G_q} - V_{\Delta W_r} V^T_{\Delta W_r} V_{\nabla G_q}.
\end{equation}

Finally, we compute Eqn. \ref{eq15} as follows:
\begin{equation}
\begin{aligned}
\label{eqf2}
f_2 = \frac{1}{q} \sum_{i=0}^{q} \text{sim}(V_{\nabla G_q}[i], (v_* - Wk_*)).
\end{aligned}
\end{equation}

\textbf{Next, we will provide a detailed explanation of the calculation formula for O-Edit+.} To ensure that the column subspaces of \(\Delta W_r\) and \(\Delta W_{[2]}\) are orthogonal, it is sufficient to ensure that the projection of \(\Delta W_{[2]}\) onto the standard orthogonal basis of the column space of \(\Delta W_r\) is zero. Similar to O-Edit, \(\Delta W_r\) is \(\Delta W_{\text{[total]}}\), and \(\nabla G_q\) is a high-rank matrix. Eqn. \ref{eq16} can be rewritten as:
\begin{equation}
\label{ortheacho+}
\begin{aligned}
\Delta W_{[2]} &= \Delta W_{[2]} - V_{\Delta W_r} V^T_{\Delta W_r} \Delta W_{[2]},\\
V_{\nabla G_q} &= V_{\nabla G_q} - V_{\Delta W_r} V^T_{\Delta W_r} V_{\nabla G_q},\\
\Delta W_{[2]} &= \Delta W_{[2]} - V_{\nabla G_q} V^T_{\nabla G_q} \Delta W_{[2]}.
\end{aligned}
\end{equation}

O-Edit and O-Edit+ are adaptations of ROME and MEMIT for sequential editing, and all experimental settings are consistent with those of ROME and MEMIT. Readers can refer to Algorithm \ref{alg1} and Algorithm \ref{alg2} for their pseudo-code.

\begin{algorithm}
\caption{Algorithm for Sequential Editing with O-Edit}
\label{alg1}
\begin{algorithmic}[1] % Line numbering
\REQUIRE  \(  \mathcal{D}_{\text{edit}} = \{(\mathcal{X}_e, \mathcal{Y}_e) \mid (x_1, y_1), \ldots, (x_T, y_T)\} \), original weight $W$, hyperparamter $r$, $q$, $\lambda_1$, $\lambda_2$, $\lambda_3$, gradient information $\nabla G$.

\ENSURE The optimal parameter $\widetilde{W}$

\FOR {$\text{Iteration} \in T$}
    \IF{$\text{Iteration} = 1$} 
        \STATE $q \leftarrow \lambda_3 \times 1$, $r \leftarrow 0$
        \STATE $\nabla G^T_{q} = U_\text{$\nabla G_q$}\Sigma_\text{$\nabla G_q$} V^T_\text{$\nabla G_q$} \leftarrow \|\nabla G_{q} - \nabla G\|_2$, subject to $\operatorname{rank}(\nabla G_{q}) = q$ \textcolor{blue}{// Obtain by calculating the SVD decomposition of $\nabla G$}
        \STATE Compute $k_*= \frac{1}{N} \sum_{j=1}^N k(s_j + x)$ (Eqn. \ref{eqk})
        \STATE Compute $v_*$ by optimizing $\mathcal{L}(v) + 0 \cdot f_1(\Delta W_{r};v) + \lambda_2 f_2(\nabla G_{q};v)$ (Eqn.\ref{eq14}) \textcolor{blue}{// Eqn. \ref{eqf1}, \ref{eqf2} for compute $f_1$ and$f_2$}.
        \STATE $\Delta W_\text{[1]} \leftarrow \Lambda(C^{-1} k_*)^T$ for ROME (Eqn.\ref{eq5}) \textcolor{blue}{// $\Delta W_1 \leftarrow Rk_*^T(C + k_*k_*^T)^{-1}$ for MEMIT (Eqn. \ref{eq8})}
        \STATE $\widetilde{W} \leftarrow W + \Delta W_\text{[1]}$ \textcolor{blue}{// Update original weight $W$ to $\widetilde{W}$}
        \STATE \textbf{Initialize} $\Delta W_\text{[total]} \leftarrow \Delta W_\text{[1]}$
    \ELSE
        \STATE $q \leftarrow \lambda_3 \times \text{Iteration}$, $r \leftarrow \text{min(1 $\times$ \text{Iteration} - 1, rank($\Delta W_\text{[total]}$)})$
        \STATE $\nabla G^T_{q} = U_\text{$\nabla G_q$}\Sigma_\text{$\nabla G_q$} V^T_\text{$\nabla G_q$} \leftarrow \|\nabla G_{q} - \nabla G\|_2$, subject to $\operatorname{rank}(\nabla G_{q}) = q$
        \STATE $\Delta W^T_{r} = U_\text{$\Delta W_r$}\Sigma_\text{$\Delta W_r$} V^T_\text{$\Delta W_r$}\leftarrow \|\Delta W_{r} - \Delta W_\text{[total]}\|_2$, subject to $\operatorname{rank}(\Delta W_{r}) = r$ \textcolor{blue}{// Actually,  $\Delta W_r = \Delta W_\text{[total]}$}
        \STATE $\nabla G_q = \nabla G_q - \Delta W_r(\Delta W^T_r \Delta W_r)^{-1} \Delta W^T_r \nabla G_q,$ \textcolor{blue}{ // Avoid knowledge conflicts, compute by Eqn.\ref{ortheacho} }
        \STATE Compute $k_*= \frac{1}{N} \sum_{j=1}^N k(s_j + x)$ (Eqn. \ref{eqk})
        \STATE Compute $v_*$ by optimizing $\mathcal{L}(v) + \lambda_1 f_1(\Delta W_{r}; v) + \lambda_2 f_2(\nabla G_{q}; v)$ (Eqn.\ref{eq14}) \textcolor{blue}{// Eqn. \ref{eqf1}, \ref{eqf2} for compute $f_1$ and$f_2$}.
        \STATE $\Delta W_\text{[Iteration]} \leftarrow \Lambda(C^{-1} k_*)^T$ for ROME (Eqn.\ref{eq5}) \textcolor{blue}{// $\Delta W_\text{[Iteration]} \leftarrow Rk_*^T(C + k_*k_*^T)^{-1}$ for MEMIT (Eqn. \ref{eq8})}
        \STATE $\widetilde{W} \leftarrow \widetilde{W} + \Delta W_\text{[Iteration]}$ \textcolor{blue}{ // Iterative update of the model weights}
        \STATE $\Delta W_\text{[total]} += \Delta W_\text{[Iteration]}$ \textcolor{blue}{ // Update the cache of $\Delta W_\text{[total]}$ }
    \ENDIF 
\ENDFOR

\RETURN update weight $\widetilde{W}$
\end{algorithmic}
\end{algorithm}

\begin{algorithm}
\caption{Algorithm for Sequential Editing with O-Edit+}
\label{alg2}
\begin{algorithmic}[1] % Line numbering
\REQUIRE  \(  \mathcal{D}_{\text{edit}} = \{(\mathcal{X}_e, \mathcal{Y}_e) \mid (x_1, y_1), \ldots, (x_T, y_T)\} \), original weight $W$, hyperparamter $r$, $q$, $\lambda_3$, gradient information $\nabla G$.

\ENSURE The optimal parameter $\widetilde{W}$

\FOR {$\text{Iteration} \in T$}
    \IF{$\text{Iteration} = 1$} 
        \STATE $q \leftarrow \lambda_3 \times 1$, $r \leftarrow 0$
        \STATE $\nabla G^T_{q} = U_\text{$\nabla G_q$}\Sigma_\text{$\nabla G_q$} V^T_\text{$\nabla G_q$} \leftarrow \|\nabla G_{q} - \nabla G\|_2$, subject to $\operatorname{rank}(\nabla G_{q}) = q$ \textcolor{blue}{// Obtain by calculating the SVD decomposition of $\nabla G$}
        \STATE Compute $k_*= \frac{1}{N} \sum_{j=1}^N k(s_j + x)$ (Eqn. \ref{eqk})
        \STATE Compute $v_*$ by optimizing $\mathcal{L}(v)$  (Eqn.\ref{eq7})
        \STATE $\Delta W_1 \leftarrow \Lambda(C^{-1} k_*)^T$ for ROME (Eqn.\ref{eq5}) \textcolor{blue}{// $\Delta W_\text{[1]} \leftarrow Rk_*^T(C + k_*k_*^T)^{-1}$ for MEMIT (Eqn. \ref{eq8})}
        \STATE  $\Delta W_\text{[1]} = \Delta W_\text{[1]} -  V_\text{$\nabla G_q$}  V^T_\text{$\nabla G_q$}\Delta W_\text{[1]}.$ (Eqn. \ref{ortheacho+})\textcolor{blue}{// Orthogonal post-processing}
        \STATE $\widetilde{W} \leftarrow W + \Delta W_\text{[1]}$ \textcolor{blue}{// Update original weight $W$ to $\widetilde{W}$}
        \STATE \textbf{Initialize} $\Delta W_\text{[total]} \leftarrow \Delta W_\text{[1]}$
    \ELSE
        \STATE $q \leftarrow \lambda_3 \times \text{Iteration}$, $r \leftarrow \text{Iteration} - 1$
        \STATE $\nabla G^T_{q} = U_\text{$\nabla G_q$}\Sigma_\text{$\nabla G_q$} V^T_\text{$\nabla G_q$} \leftarrow \|\nabla G_{q} - \nabla G\|_2$, subject to $\operatorname{rank}(\nabla G_{q}) = q$ 
        \STATE $\Delta W^T_{r} = U_\text{$\Delta W_r$}\Sigma_\text{$\Delta W_r$} V^T_\text{$\Delta W_r$}\leftarrow \|\Delta W_{r} - \Delta W_\text{[total]}\|_2$, subject to $\operatorname{rank}(\Delta W_{r}) = r$ \textcolor{blue}{// Actually,  $\Delta W_r = \Delta W_\text{[total]}$}
        \STATE Compute $k_*= \frac{1}{N} \sum_{j=1}^N k(s_j + x)$ (Eqn. \ref{eqk})
        \STATE Compute $v_*$ by optimizing $\mathcal{L}(v)$  (Eqn.\ref{eq7})
        \STATE $\Delta W_\text{[Iteration]} \leftarrow \Lambda(C^{-1} k_*)^T$ for ROME (Eqn.\ref{eq5}) \textcolor{blue}{// $\Delta W_\text{Iteration} \leftarrow Rk_*^T(C + k_*k_*^T)^{-1}$ for MEMIT (Eqn. \ref{eq8})}
        \STATE $\Delta W_\text{[Iteration]} = \Delta W_\text{[Iteration]} - V_{\Delta W_r} V^T_{\Delta W_r}\Delta W_\text{[Iteration]}$ \\
        $V_\text{$\nabla G_q$} = V_\text{$\nabla G_q$} - V_\text{$\Delta W_r$} V^T_\text{$\Delta W_r$}V_\text{$\nabla G_q$}$, (Eqn. \ref{ortheacho+}) 
        \textcolor{blue}{ // Orthogonal post-processing}\\
        $\Delta W_\text{[Iteration]} = \Delta W_\text{[Iteration]} -  V_\text{$\nabla G_q$} V^T_\text{$\nabla G_q$}\Delta W_\text{[Iteration]}$  \\

        \STATE $\widetilde{W} \leftarrow \widetilde{W} + \Delta W_\text{[Iteration]}$  \textcolor{blue}{ // Iterative update of the model weights}
        \STATE $\Delta W_\text{[total]} += \Delta W_\text{[Iteration]}$ \textcolor{blue}{ // Update the cache of $\Delta W_\text{[total]}$ }
    \ENDIF 
\ENDFOR

\RETURN update weight $\widetilde{W}$
\end{algorithmic}
\end{algorithm}

\section{Experimental Setup}
\subsection[How to choose an appropriate gradient]{How to choose an appropriate $\nabla G_q$}

\label{G}

The core of our method lies in capturing the update direction of implicit knowledge within the model. Theoretically, if we view the model as a knowledge base \citep{petroni2019languagemodelsknowledgebases}, the update direction should align with the gradient direction in which the model continues to learn from this knowledge. Thus, selecting the appropriate knowledge base is crucial for determining the model's update gradient. We explored the following methods:

\begin{itemize}
    \item We selected 100,000 pieces of unrelated knowledge from COUNTERFACT, which are outside the experimental test samples. This set, referred to as ``locality\_prompt'' in Figure \ref{cf}, serves as the expected gradient direction.

    \item We utilized the knowledge employed by \citep{meng2023locatingeditingfactualassociations}, which successfully identified how knowledge is stored within the model.

    \item For comparison, we randomly generated 100,000 text samples using ASCII codes.

    \item We also used Wikipedia as a knowledge source, as it is commonly chosen for pre-training in large language models (LLMs).
\end{itemize}

The experimental results are presented in Table \ref{corpus}. We maintained consistency in the parameters related to \(\Delta W_\text{[total]}\) across experiments, with the only variable being the source of the \(\nabla G\) corpus. Randomly generated text yielded the poorest performance, while the "locality\_prompt" from COUNTERFACT achieved the second-best results, only surpassed by Wikipedia, which produced the best outcomes. These results also serve as reverse validation that the implicit knowledge within the model is embedded in its pre-training data.

\begin{table}[H]  
\vspace{0.5cm}  
    \caption{\textbf{Different corpus results for COUNTERFACT.} $T$: Num Edits.}  
    \centering  
    \resizebox{\linewidth}{!}{  
    \begin{tabular}{lccc|c|ccc|c|ccc|c|ccc|c}  
    \toprule  
    \multirow{3}{*}{\textbf{MEMIT}} & \multicolumn{16}{c}{\textbf{COUNTERFACT}} \\
    \cmidrule(lr){2-17}  
     & \multicolumn{4}{c|}{$T=200$} & \multicolumn{4}{c|}{$T=500$} & \multicolumn{4}{c|}{$T=1000$} & \multicolumn{4}{c}{$T=1500$} \\
     \midrule  
     & Rel. & Gen. & Loc. & Avg. & Rel. & Gen. & Loc. & Avg.  & Rel. & Gen. & Loc. & Avg. & Rel. & Gen. & Loc. & Avg. \\
     \midrule  
     \multicolumn{17}{c}{\texttt{\textbf{Mistral-7B}}} \\
     \midrule  
    \cmidrule(lr){1-17}  
     \qquad \textbf{Corpus\ding{202}} & 0.89 & 0.62 & 0.74 & 0.75 & 0.79 & 0.55 & \textbf{0.61} & 0.65 & 0.64 & 0.37 & 0.52 & 0.54 & 0.57 & 0.39 & 0.51 & 0.49 \\
     \qquad \textbf{Corpus\ding{203}}  & \textbf{0.90} & \textbf{0.62} & 0.73 & 0.75 & 0.76 & 0.53 & 0.60 & 0.63 & 0.62 & 0.33 & 0.50 & 0.48 & 0.54 & 0.36 & 0.49 & 0.46 \\
     \qquad \textbf{Corpus\ding{204}}  & 0.86 & 0.60 & 0.73 & 0.73 & 0.74 & 0.51 & 0.56 & 0.60 & 0.59 & 0.32 & 0.44 & 0.45 & 0.57 & 0.33 & 0.46 & 0.45 \\
     \rowcolor{red!8}  
     \qquad \textbf{Corpus\ding{205}} & 0.89 & 0.61 &\textbf{ 0.78} & \textbf{0.76} & \textbf{0.81} & \textbf{0.55} & 0.60 & \textbf{0.65} & \textbf{0.68} & \textbf{0.39} & \textbf{0.55} & \textbf{0.54} & \textbf{0.61} & \textbf{0.42} & \textbf{0.53} & \textbf{0.52} \\
      \cmidrule(lr){1-17}  
     \multicolumn{17}{c}{\texttt{\textbf{Llama3-8B}}} \\
     \midrule  
    \cmidrule(lr){1-17}  
     \qquad \textbf{Corpus\ding{202}} & 0.88 & 0.47 & 0.65 & 0.67 & \textbf{0.85} & 0.48 & 0.36 & 0.56 & 0.79 & 0.47 & 0.29 & 0.52 & 0.77 & 0.46 & 0.26 & 0.49 \\
     \qquad \textbf{Corpus\ding{203}} & \textbf{0.91} & 0.48 & 0.66 & 0.68 & 0.85 & 0.50 & 0.40 & 0.58 & 0.79 & 0.47 & 0.30 & 0.52 & 0.74 & 0.44 & 0.27 & 0.48 \\
     \qquad \textbf{Corpus\ding{204}} & 0.85 & 0.41 & 0.63 & 0.63 & 0.83 & 0.45 & 0.31 & 0.53 & 0.74 & 0.41 & 0.24 & 0.46 & 0.70 & 0.35 & 0.19 & 0.41 \\
     \rowcolor{red!8}  
     \qquad \textbf{Corpus\ding{205}} & 0.88 & \textbf{0.53} & \textbf{0.76} & \textbf{0.72} & 0.84 & \textbf{0.51} & \textbf{0.45} & \textbf{0.60} & \textbf{0.81} & \textbf{0.50} & \textbf{0.31} & \textbf{0.54} & \textbf{0.79} & \textbf{0.44} & \textbf{0.28} & \textbf{0.50} \\
     \bottomrule  
    \end{tabular}  
    }  
    \label{corpus}  
    \vspace{-0.4cm}  
\end{table}

\subsection{Baseline Editing Methods}
\label{baseline}
We selected five popular model editing methods as baselines:

\begin{itemize}  
    \item \textbf{ROME} \citep{meng2023locatingeditingfactualassociations} has been previously discussed. In this experiment, we edit the 8th layer, which is regarded as a crucial location for knowledge storage. We utilize second moment statistics \( C \propto E[kk^T] \) computed from more than 100,000 samples of hidden states \( k \) derived from tokens sampled across all Wikipedia text in context.  

    \item \textbf{MEMIT} \citep{meng2023masseditingmemorytransformer}—the detailed computation process can be found in Appendix \ref{memit}. We set \( \lambda = 15,000 \) to balance the knowledge in the model with the knowledge required for editing. Other settings are consistent with those in ROME.

    \item \textbf{R-Edit} \citep{gupta2024rebuildingromeresolving} attributes the suboptimal performance of ROME and MEMIT to the inadequacy of the calculated \( k_* \) in representing the subject of the queried knowledge. R-Edit enhances the calculation of \( k_* \) in Eqs. \ref{eq5} and \ref{eq8} to address this issue.

    \item \textbf{WilKE} \citep{hu2024wilkewiselayerknowledgeeditor} argues that different types of knowledge should be distributed across various layers. For each piece of knowledge edited, WilKE first determines the optimal layer for editing and then applies either ROME or MEMIT to perform the edit. Due to the time and computational cost of finding the optimal layer, we restrict the editable layers in this paper to \( l = \{5, 6, 7, 8, 9, 10\} \).

    \item \textbf{PRUNE} \citep{ma2024perturbationrestrainedsequentialmodelediting} suggests that the key factor influencing sequential editing performance is the condition number of the matrix. PRUNE scales the singular values in \( \Delta W_\text{total} \) that exceed the maximum singular value of the original model, ensuring that no singular value surpasses a specified threshold. We adhere to the experimental setup outlined in \cite{ma2024perturbationrestrainedsequentialmodelediting} and scale the larger singular values using the following method:  
    \[
    F(\hat{\sigma}_i) = \log_{1.2}(\hat{\sigma}_i) - \log_{1.2}(\max{\{\sigma_i\}}) + \max{\{\sigma_i\}}.  
    \]
\end{itemize}

\begin{table}[H]
\vspace{-0.5cm}
    \caption{\textbf{Different orthogonal method results for COUNTERFACT.} $T$: Num Edits.}
    \centering
    % \setstretch{1.2}
    \resizebox{\linewidth}{!}{
    \begin{tabular}{lccc|c|ccc|c|ccc|c|ccc|c}
    \toprule
    \multirow{3}{*}{\textbf{Method}} & \multicolumn{16}{c}{\textbf{COUNTERFACT}} \\
     \cmidrule(lr){2-17}
     & \multicolumn{4}{c|}{$T=200$} & \multicolumn{4}{c|}{$T=500$} & \multicolumn{4}{c|}{$T=1000$} & \multicolumn{4}{c}{$T=1500$} \\
     \midrule
     & Rel. & Gen. & Loc. & Avg. & Rel. & Gen. & Loc. & Avg.  & Rel. & Gen. & Loc. & Avg. & Rel. & Gen. & Loc. & Avg. \\
     \midrule
     \multicolumn{17}{c}{\texttt{\textbf{Mistral-7B}}} \\
     \midrule
     \cmidrule(lr){1-17}
     \textbf{MEMIT} & \textbf{0.93} & \textbf{0.67} & 0.41 & 0.67 & 0.50 & 0.35 & 0.10 & 0.32 & 0.28 & 0.10 & 0.06 & 0.15 & 0.19 & 0.06 & 0.05 & 0.10 \\
     \cmidrule(lr){1-17}
     \qquad \ding{202}\textbf{Only}$\Delta W_\text{[total]}$ & 0.91 & 0.54 &0.77 & 0.74 & 0.79 & 0.53 & 0.55 & 0.62 & 0.61 & 0.37 & 0.51 & 0.50 & 0.55 & 0.37 & 0.46 & 0.46\\
     \qquad \ding{203}\textbf{Only}$\nabla G$ & 0.89 & 0.55 & 0.74 & 0.72 & 0.76 & 0.50 & 0.51 & 0.59 & 0.57 & 0.34 & 0.49 & 0.47 & 0.44 & 0.27 & 0.24 & 0.32 \\
     \qquad \ding{204}\textbf{Without Eqn.\ref{eq13}(\ref{ortheacho})} & 0.89 & 0.59 &0.77 & 0.75 & 0.78 & 0.56 & 0.56 & 0.63 & 0.58 & 0.37 & 0.52 & 0.49 & 0.49 & 0.36 & 0.49 & 0.44 \\
      \rowcolor{red!8}
     \qquad \textbf{O-Edit+} & 0.89 & 0.61 &\textbf{ 0.78} & \textbf{0.76} & \textbf{0.81} & \textbf{0.55} & \textbf{0.60} & \textbf{0.65} & \textbf{0.68} & \textbf{0.39} & \textbf{0.55} & \textbf{0.54} & \textbf{0.61} & \textbf{0.42} & \textbf{0.53} & \textbf{0.52} \\
     \midrule
     \multicolumn{17}{c}{\texttt{\textbf{Llama3-8B}}} \\
     \midrule
     \cmidrule(lr){1-17}
     \textbf{MEMIT} & 0.85 & 0.51 & 0.22 & 0.52 & 0.50 & 0.35 & 0.10 & 0.32 & 0.28 & 0.10 & 0.05 & 0.14 & 0.18 & 0.06 & 0.05 & 0.10 \\
     \cmidrule(lr){1-17}
     \qquad \ding{202}\textbf{Only}$\Delta W_\text{[total]}$ & \textbf{0.91} & 0.49 & 0.65 & 0.68 & 0.87 &\textbf{ 0.54} & 0.36 & 0.59 & 0.78 & 0.45 & 0.28 & 0.50 & 0.74 & 0.41 & 0.25 & 0.47 \\
     \qquad \ding{203}\textbf{Only}$\nabla G$ & 0.87 & 0.49 & 0.62 & 0.66 & 0.77 & 0.41 & 0.32 & 0.50 & 0.64 & 0.32 & 0.28 & 0.41 & 0.55 & 0.28 & 0.22 & 0.35 \\ 
     \qquad \ding{204}\textbf{Without Eqn.\ref{eq13}(\ref{ortheacho})} & 0.88 & 0.48 & 0.65 & 0.67 & \textbf{0.87} & 0.50 & 0.41 & 0.60 & 0.78 & 0.46 & 0.32 & 0.52 & 0.67 & 0.39 & 0.28 & 0.43 \\
     \rowcolor{red!8}
     \qquad\textbf{O-Edit+} & 0.88 & \textbf{0.53} & \textbf{0.76} & \textbf{0.72} & 0.84 & 0.51 & \textbf{0.45} & \textbf{0.60} & \textbf{0.81} & \textbf{0.50} & \textbf{0.31} & \textbf{0.54} & \textbf{0.79} & \textbf{0.44} & \textbf{0.28} & \textbf{0.50} \\
     \bottomrule
    \end{tabular}
    }
    \label{ablation1}
    \vspace{-0.4cm}
\end{table}

\subsection{Experiments Compute Resources time and hyperparameters}
\label{oeditsettings}

We used NVIDIA A100 40GB GPUs for the experiments. For Mistral-7B and LLaMA3-8B, ROME and MEMIT occupy about 35GB of memory and take about 2.5 hours to run 1500 edits. For O-Edit, running 1500 edits requires about 6 hours, while O-Edit+ takes about 5 hours. 

For all experimental settings of O-Edit, we set \( \lambda_1 \) and \( \lambda_2 = 50 \). For O-Edit+, we set \( \lambda_3 = 2 \) for Mistral-7B and \( \lambda_3 = 1 \) for LLaMA-8B in MEMIT; \( \lambda_3 = 2.5 \) for both Mistral-7B and LLaMA3-8B in ROME. In the next Section \ref{Ablation}, we conducted detailed ablation experiments and parameter selection experiments to further analyze the impact of hyperparameters on editing performance.

Another potential issue arises when \( q \) exceeds the dimensions of the model (\( \min(d, d_m) \))\footnote{The dimension of \( W_{proj} \) in both Mistral-7B and LLaMA3-8B is \( \mathbb{R}^{4096 \times 14336} \).}. In this paper, we have considered 1500 edits. When the number of required edits exceeds this amount, \( q \) can be constrained by setting it below a certain threshold to ensure the feasibility of performing additional edits. A smaller threshold for \( q \) typically results in more effective edits, while a larger threshold tends to preserve the model's ability to retain unrelated knowledge. However, in general, increasing the number of edits tends to cause greater degradation in the model's performance.

\subsection{Ablation Experiments}
\label{Ablation}
First, we wanted to see if both \( \Delta W_\text{[total]} \) and \( \nabla G \) contributed effectively. We set up three baselines: \ding{202} using only \( \Delta W_\text{[total]} \); \ding{203} using only \( \nabla G \); and \ding{204} using both \( \Delta W_\text{[total]} \) and \( \nabla G \) without orthogonal processing for \( \nabla G \) according to Eq.\ref{eq13}(\ref{ortheacho}). The results are shown in Table \ref{ablation1}. We observed that while using either \( \Delta W_\text{[total]} \) or \( \nabla G \) alone yielded better results than the original method, their performance was still inferior to using both together. The lack of orthogonalization for \( \nabla G \) led to knowledge conflicts within the model, resulting in inferior performance compared to O-Edit+.

Next, we compared the effects of different hyperparameter selections on editing performance between O-Edit and O-Edit+, as shown in Tables \ref{hpyeroedit+} and \ref{hpyeroedit}. In O-Edit+, two noteworthy phenomena were observed. First, MEMIT’s \( \lambda_3 \) is smaller than that of ROME due to ROME's stronger constraints, which can degrade the performance of unrelated knowledge (\textbf{Loc.}) during sequential editing. Consequently, we opted for a larger \( \lambda_3 = 2.5 \) to mitigate ROME’s influence. Second, while a smaller \( \lambda_3 \) improves performance with MEMIT, it still negatively impacts unrelated knowledge, and a larger \( \lambda_3 \) affects the editing effect (\textbf{Rel., Gen.}). Therefore, selecting an appropriately sized \( \lambda_3 \) is crucial for optimal overall editing performance.

In the O-Edit setting, we compared the editing performance under four different settings. The results showed that stronger constraints led to better outcomes, as \( \lambda_1 \) and \( \lambda_2 \) effectively controlled the correlation between different edits. Larger \( \lambda_1 \) values resulted in smaller correlations between edits, while larger \( \lambda_2 \) values reduced the correlation between edited and implicit knowledge within the model.

\begin{table}[t]

    \caption{\textbf{Hpyerparameter selection results for O-Edit+.} $T$: Num Edits.}
    \centering
    % \setstretch{1.2}
    \resizebox{\linewidth}{!}{
    \begin{tabular}{lccc|c|ccc|c|ccc|c|ccc|c}
    \toprule
    \multirow{3}{*}{\textbf{Method}} & \multicolumn{16}{c}{\textbf{COUNTERFACT}} \\
     \cmidrule(lr){2-17}
     & \multicolumn{4}{c|}{$T=200$} & \multicolumn{4}{c|}{$T=500$} & \multicolumn{4}{c|}{$T=1000$} & \multicolumn{4}{c}{$T=1500$} \\
     \midrule
     & Rel. & Gen. & Loc. & Avg. & Rel. & Gen. & Loc. & Avg.  & Rel. & Gen. & Loc. & Avg. & Rel. & Gen. & Loc. & Avg. \\
     \midrule
     \multicolumn{17}{c}{\texttt{\textbf{Mistral-7B}}} \\
     \midrule
    \textbf{ROME} & 0.72 & 0.53 & 0.31 & 0.52 & 0.30 & 0.18 & 0.14 & 0.21 & 0.28 & 0.10 & 0.06 & 0.15 & 0.27 & 0.13 & 0.05 & 0.13 \\
    \cmidrule(lr){1-17}

     \qquad $\lambda_3 = 1$ & 0.92 & \textbf{0.50} & 0.73 & 0.71 & 0.60 & 0.34 & 0.34 & 0.42 & 0.38 &  0.13 & 0.17 & 0.22 & 0.35 & 0.17 & 0.10 & 0.20 \\
     \qquad $\lambda_3 = 2$ & \textbf{0.95} & 0.47 & 0.73 & 0.71 & 0.64 & 0.34 & 0.40 & 0.46 &  0.43 & 0.16 & 0.21 & 0.26 & 0.37 & 0.19 & 0.17 &0.24 \\
     \rowcolor{red!8}
     \qquad $\lambda_3 = 2.5$ & 0.94 & 0.47 & \textbf{0.76} & \textbf{0.72} & \textbf{0.65} & \textbf{0.38} & \textbf{0.41} & \textbf{0.48} & \textbf{0.49} & \textbf{0.21} & \textbf{0.29} & \textbf{0.33} & \textbf{0.41} & \textbf{0.21} & \textbf{0.24} & \textbf{0.29} \\
     \cmidrule(lr){1-17}
     \textbf{MEMIT} & \textbf{0.93} & \textbf{0.67} & 0.41 & 0.67 & 0.50 & 0.35 & 0.10 & 0.32 & 0.28 & 0.10 & 0.06 & 0.15 & 0.19 & 0.06 & 0.05 & 0.10 \\
     \cmidrule(lr){1-17}
     \qquad $\lambda_3 =1$ & 0.88 & 0.53 & 0.76 & 0.72 & 0.81 & 0.47 & 0.56 & 0.61 & 0.70 & 0.38 & 0.48 & 0.52 & 0.60 & 0.30 & 0.44 & 0.44 \\
      \qquad $\lambda_3 =2.5$ & 0.84 & 0.50 & \textbf{0.84} & 0.73 & 0.77 & 0.40 & \textbf{0.62} & 0.59 & 0.62 & 0.31 & \textbf{0.61} & 0.51 & 0.55 & 0.23 & \textbf{0.56} & 0.44 \\
      \rowcolor{red!8}
     \qquad $\lambda_3 =2$ & 0.89 & 0.61 & 0.78 & \textbf{0.76} & \textbf{0.81} & \textbf{0.55} & 0.60 & \textbf{0.65} & \textbf{0.68} & \textbf{0.39} & 0.55 & \textbf{0.54} & \textbf{0.61} & \textbf{0.42} & 0.53 & \textbf{0.52} \\
     \midrule
     \multicolumn{17}{c}{\texttt{\textbf{Llama3-8B}}} \\
     \midrule
     \textbf{ROME} & 0.75 & 0.48 & 0.14 & 0.46 & 0.69 & 0.45 & 0.05 & 0.40 & 0.75 & 0.46 & 0.02 & 0.41 & 0.47 & 0.28 & 0.02 & 0.31 \\
     \cmidrule(lr){1-17}
     \qquad $\lambda_3 = 1$ & \textbf{0.88} & 0.47 & 0.30 & 0.55 & \textbf{0.84} & 0.47 & 0.10 & 0.47 & 0.78 & 0.48 & 0.07 & 0.44 & 0.76 & 0.34 & 0.07 & 0.39 \\
     \qquad $\lambda_3 =2$ & 0.85 & 0.50 & \textbf{0.38} & 0.58 & 0.80 & 0.51 & 0.13 & 0.48 & \textbf{0.87} & 0.46 & 0.09 & 0.47 & 0.84 & 0.39 & 0.09 & 0.44 \\
     \rowcolor{red!8}
     \qquad $\lambda_3 =2.5$  & 0.86 & \textbf{0.61} & 0.37 & \textbf{0.61} & 0.81 & \textbf{0.52} & \textbf{0.24} & \textbf{0.52} & 0.86 & \textbf{0.49} & \textbf{0.19} & \textbf{0.51} & \textbf{0.87} & \textbf{0.50} & \textbf{0.13} & \textbf{0.50} \\
    \cmidrule(lr){1-17}
     \textbf{MEMIT} & 0.85 & 0.51 & 0.22 & 0.52 & 0.50 & 0.35 & 0.10 & 0.32 & 0.28 & 0.10 & 0.05 & 0.14 & 0.18 & 0.06 & 0.05 & 0.10 \\
     \cmidrule(lr){1-17}
     \qquad $\lambda_3 = 0.5$ & 0.88 & 0.47 & 0.61 & 0.65 & 0.84 & 0.47 & 0.38 & 0.56 & 0.78 & 0.48 & 0.30 & 0.52 & 0.76 & \textbf{0.46} & 0.25 & 0.49 \\
     \qquad $\lambda_3 =2$ & 0.86 & 0.49 & 0.66 & 0.67 & 0.84 & \textbf{0.52} & 0.42 & 0.59 & 0.78 & 0.46 & \textbf{0.33} & 0.52 & 0.73 & 0.43 & \textbf{0.30} & 0.49 \\
     \rowcolor{red!8}
     \qquad $\lambda_3 = 1$ & \textbf{0.88} & \textbf{0.53} & \textbf{0.76} & \textbf{0.72} & \textbf{0.84} & 0.51 & \textbf{0.45} & \textbf{0.60} & \textbf{0.81} & \textbf{0.50} & 0.31 & \textbf{0.54} & \textbf{0.79} & 0.44 & 0.28 & \textbf{0.50} \\
     \bottomrule
    \end{tabular}
    }
    \label{hpyeroedit+}
    \vspace{-0.4cm}
\end{table}

\subsection{Editing Datasets}
\label{Editing Datasets}

\begin{figure}[ht]  
    \centering  
    \includegraphics[width=\textwidth]{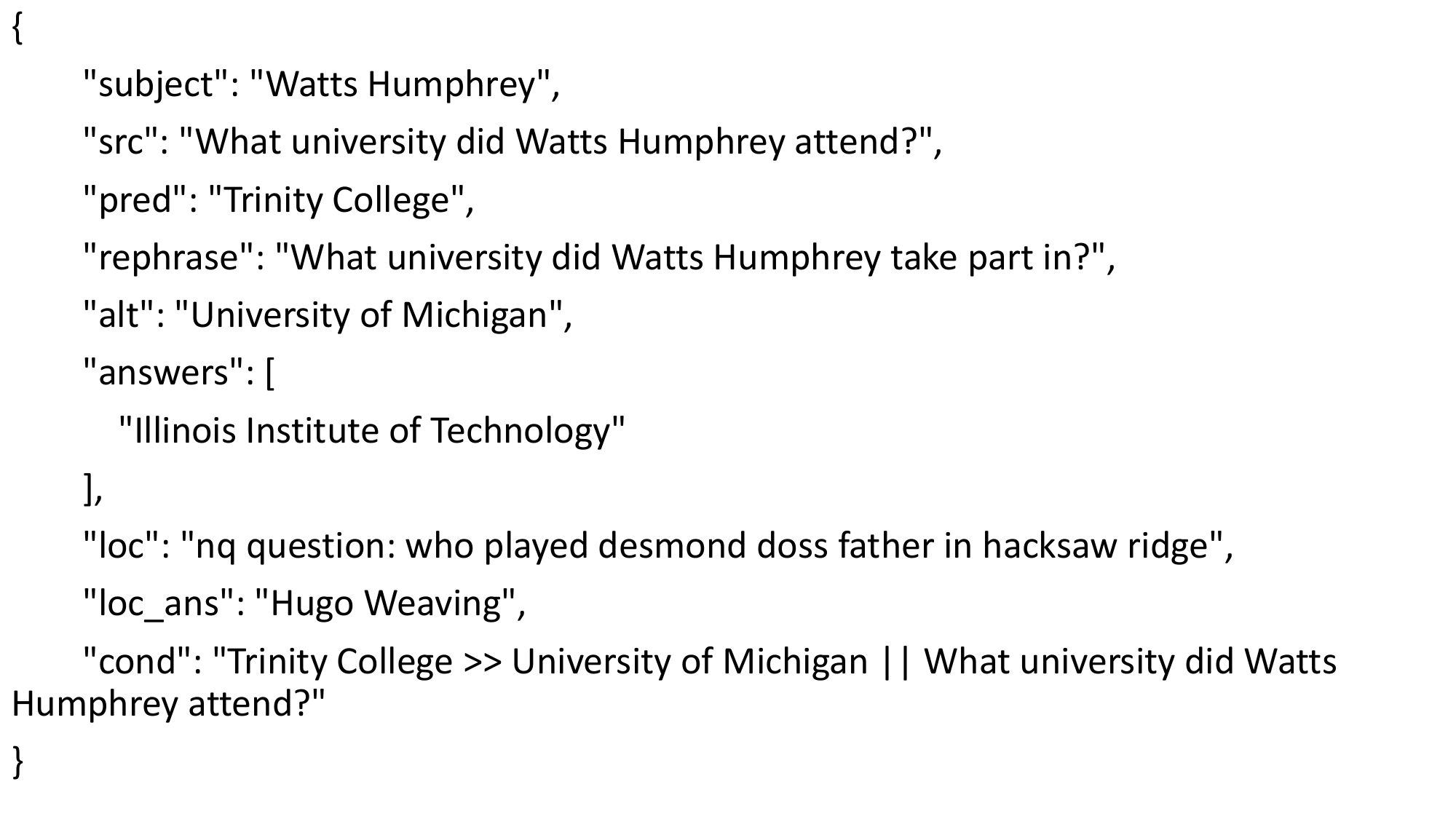}  
    \caption{Sample of ZsRE Dataset}  
    \label{zsre}  
\end{figure}

\begin{figure}[ht]   
    \centering  
    \includegraphics[width=\textwidth]{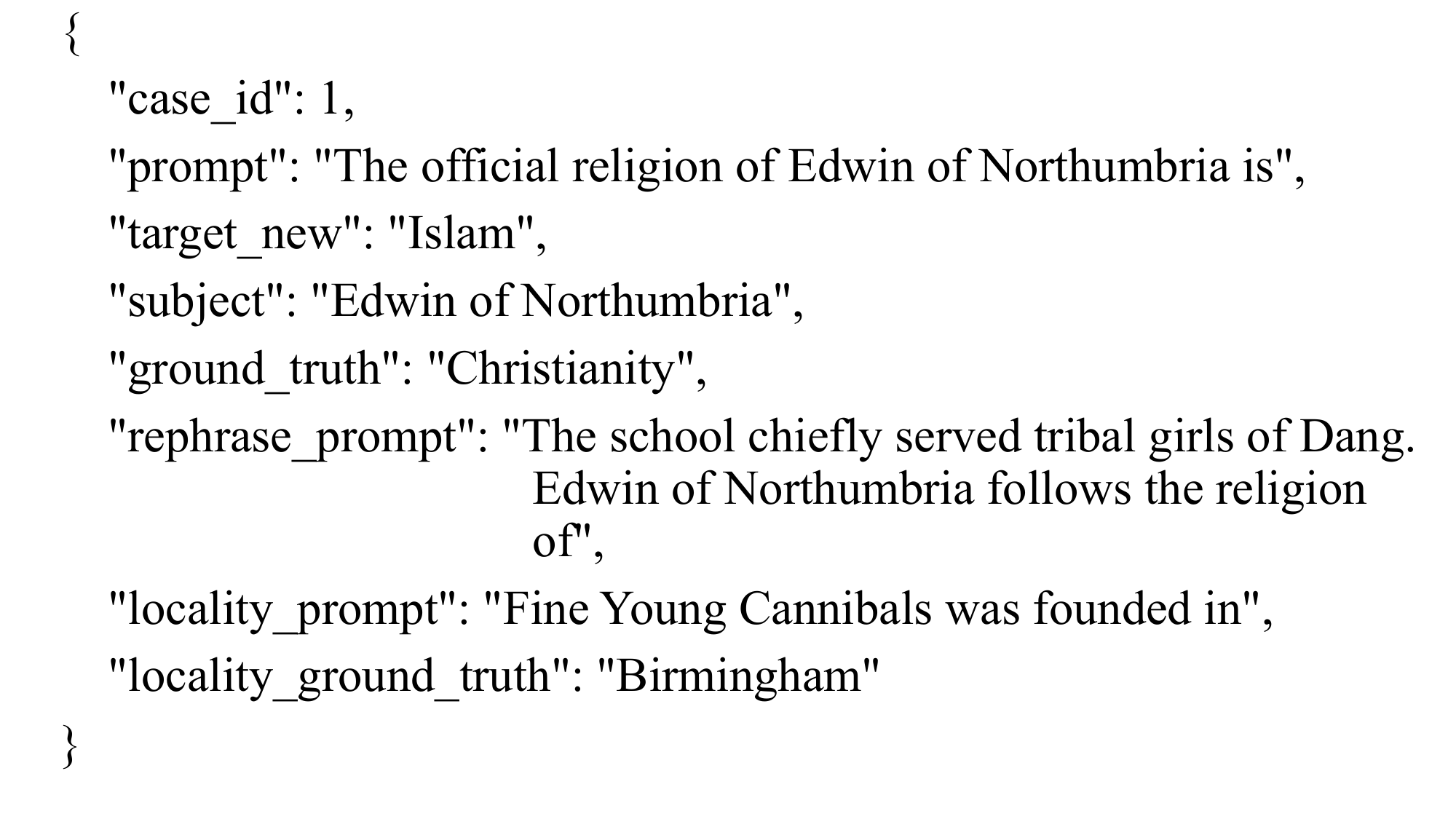}  
    \caption{Sample of COUNTERFACT Dataset}  
    \label{cf}  
\end{figure}

\begin{itemize}

\item \textbf{ZsRE} question answering task \citep{levy-etal-2017-zero} was first used for factual knowledge evaluation by \citep{decao2021editingfactualknowledgelanguage}, later being extended and adopted by \citep{mitchell2022fastmodeleditingscale}. We conduct the experiment using the version provided by \citep{yao2023editinglargelanguagemodels} in EasyEdit\footnote{\url{https://github.com/zjunlp/EasyEdit}}. Figure \ref{zsre} shows examples from ZsRE.

\item \textbf{COUNTERFACT} is designed to enable distinction between superficial changes in model word choices from specific and generalized changes in underlying factual knowledge. Figure \ref{cf} shows examples from COUNTERFACT.
\end{itemize}

\begin{table}[H]
\vspace{-0.5cm}
    \caption{\textbf{Hpyerparameter selection results for O-Edit.} $T$: Num Edits.}
    \centering
    % \setstretch{1.2}
    \resizebox{\linewidth}{!}{
    \begin{tabular}{lccc|c|ccc|c|ccc|c|ccc|c}
    \toprule
    \multirow{3}{*}{\textbf{Method}} & \multicolumn{16}{c}{\textbf{COUNTERFACT}} \\
     \cmidrule(lr){2-17}
     & \multicolumn{4}{c|}{$T=200$} & \multicolumn{4}{c|}{$T=500$} & \multicolumn{4}{c|}{$T=1000$} & \multicolumn{4}{c}{$T=1500$} \\
     \midrule
     & Rel. & Gen. & Loc. & Avg. & Rel. & Gen. & Loc. & Avg.  & Rel. & Gen. & Loc. & Avg. & Rel. & Gen. & Loc. & Avg. \\
    \cmidrule(lr){1-17}
     \midrule
     \multicolumn{17}{c}{\texttt{\textbf{Mistral-7B}}} \\
     \midrule
     \textbf{MEMIT} & 0.93 & \textbf{0.67} & 0.41 & 0.67 & 0.50 & 0.35 & 0.10 & 0.32 & 0.28 & 0.10 & 0.06 & 0.15 & 0.19 & 0.06 & 0.05 & 0.10 \\
     \cmidrule(lr){1-17}
     \qquad $\lambda_1, \lambda_2 =1$ & \textbf{0.95} & 0.66 & 0.52 & 0.68 & 0.74 & 0.36 & 0.29 & 0.46 & 0.40 & 0.24 & 0.11 & 0.25 & 0.39 & 0.19 & 0.08 & 0.22\\
     \qquad $\lambda_1, \lambda_2 =10$ & 0.93 & 0.62 & 0.54 & 0.70 & \textbf{0.89} & 0.50 & 0.35 & 0.58 & 0.54 & 0.26 & 0.18 & 0.31 & 0.45 & 0.22 & 0.10 & 0.26 \\
     \qquad $\lambda_1, \lambda_2 =20$ & 0.93 & 0.53 & 0.62 & 0.69 & 0.87 & 0.52 & 0.36 & 0.58 & 0.64 & 0.39 & 0.24 & 0.42 & 0.47 & 0.26 & 0.13 & 0.29 \\
      \rowcolor{green!8}
      \qquad $\lambda_1, \lambda_2 =50$ & 0.93 & 0.55 & \textbf{0.65} & \textbf{0.71} & 0.86 &\textbf{ 0.53} & \textbf{0.45} & \textbf{0.61} & \textbf{0.72} & \textbf{0.47} &\textbf{0.34} & \textbf{0.51} & \textbf{0.51} & \textbf{0.33} & \textbf{0.18} & \textbf{0.34} \\

     \midrule
     \multicolumn{17}{c}{\texttt{\textbf{Llama3-8B}}} \\
     \midrule
     \textbf{MEMIT} & 0.85 & 0.51 & 0.22 & 0.52 & 0.50 & 0.35 & 0.10 & 0.32 & 0.28 & 0.10 & 0.05 & 0.14 & 0.18 & 0.06 & 0.05 & 0.10 \\
     \cmidrule(lr){1-17}
     \qquad $\lambda_1, \lambda_2 =1$ & 0.96 & 0.52 & 0.43 & 0.63 & 0.83 & 0.59 & 0.16 & 0.52 & 0.62 & 0.49 & 0.08 & 0.40 & 0.35 & 0.27 & 0.08 & 0.23 \\
     \qquad $\lambda_1, \lambda_2 =10$ & \textbf{0.97} & 0.47 & 0.53 & 0.65 & 0.90 & 0.55 & 0.35 & 0.60 & 0.72 & 0.51 & 0.18 & 0.47 & 0.45 & 0.33 &0.10 & 0.29 \\
     \qquad $\lambda_1, \lambda_2 =20$ & 0.96 & 0.42 & 0.57 & 0.65 & \textbf{0.90} & \textbf{0.54} & 0.41 & 0.61 & \textbf{0.75} & \textbf{0.52} & 0.22 & 0.49 & 0.45 & 0.35 & 0.15 & 0.31 \\
     \rowcolor{green!8}
     \qquad $\lambda_1, \lambda_2 =50$ & 0.93 & \textbf{0.55} & \textbf{0.64} & \textbf{0.71} & 0.86 & 0.53 & \textbf{0.44} & \textbf{0.61} & 0.72 & 0.47 & \textbf{0.33} & \textbf{0.51} & \textbf{0.55} & \textbf{0.40} & \textbf{0.27} & \textbf{0.41}  \\
     \bottomrule
    \end{tabular}
    }
    \label{hpyeroedit}
    \vspace{-0.4cm}
\end{table}

\subsection{DOWNSTREAM TASKS SETTINGS}
\label{downstream}
To explore the side effects of sequential model editing on the general abilities of LLMs, four
representative tasks with corresponding datasets were adopted for assessment, including: \textbf{Commonsense Reasoning} on the \textbf{SIQA} \citep{sap2019socialiqacommonsensereasoningsocial}, which is a benchmark for testing social commonsense intelligence. \textbf{Content Analysis} on the \textbf{LAMBADA} \citep{paperno2016lambadadatasetwordprediction}, which is a collection of narrative paragraphs that requires computational models to track information across a broader discourse. \textbf{Question Answering} on the \textbf{CommonsenseQA} \citep{talmor2019commonsenseqaquestionansweringchallenge}, it requires the model be capable of making reasonable inferences under given common sense conditions. \textbf{MATH} on the \textbf{GSM8K} \citep{cobbe2021trainingverifierssolvemath}, a dataset of 8.5K high-quality linguistically diverse grade school math word problems. The prompts for each downstream task were illustrated in Table \ref{zero-shot}. We utilized OpenCompass\footnote{\url{https://github.com/open-compass/opencompass}} to conduct our evaluations.

\begin{table}[H]
\centering
\caption{The prompts to LLMs for evaluating their zero-shot performance on these general tasks.}
\label{zero-shot}
\begin{tabular}{m{3cm} m{9cm}}
\toprule
\textbf{Task} & \textbf{Prompt} \\
\midrule
SIQA & 
\parbox{9cm}{
  prompt=
  ``\{question\} A. \{A\} B. \{B\} C. \{C\} Answer:''
} \\
\midrule
LAMBADA & 
\parbox{9cm}{
  prompt=``Please complete the following sentence: \{sentence\}''
} \\
\midrule
CommonsenseQA & 
\parbox{9cm}{
  prompt=
  ``\{question\} A. \{A\} B. \{B\} C. \{C\} D. \{D\} E. \{E\} Answer:''
} \\
\midrule
GSM8K & 
\parbox{9cm}{
  prompt = ``
  Question: \{question\}
  Let's think step by step.
  Answer:''
} \\
\bottomrule
\end{tabular}
\end{table}

\begin{table}[H]
\vspace{-0.5cm}
    \caption{\textbf{The results of different method with similar \(\|\Delta W_\text{[total]} \|_2\).} $T$: Num Edits.}
    \centering
    % \setstretch{1.2}
    \resizebox{\linewidth}{!}{
    \begin{tabular}{lccc|c|ccc|c|ccc|c|ccc|c}
    \toprule
    \multirow{3}{*}{\textbf{Method}} & \multicolumn{16}{c}{\textbf{COUNTERFACT}} \\
     \cmidrule(lr){2-17}
     & \multicolumn{4}{c|}{$T=200$} & \multicolumn{4}{c|}{$T=500$} & \multicolumn{4}{c|}{$T=1000$} & \multicolumn{4}{c}{$T=1500$} \\
     \midrule
     & Rel. & Gen. & Loc. & Avg. & Rel. & Gen. & Loc. & Avg.  & Rel. & Gen. & Loc. & Avg. & Rel. & Gen. & Loc. & Avg. \\
     \midrule
     \multicolumn{17}{c}{\texttt{\textbf{Mistral-7B}}} \\
     \midrule

     \textbf{MEMIT} & \textbf{0.93} & 0.67 & 0.41 & 0.67 & 0.50 & 0.35 & 0.10 & 0.32 & 0.28 & 0.10 & 0.06 & 0.15 & 0.19 & 0.06 & 0.05 & 0.10 \\
     \cmidrule(lr){1-17}
     \qquad Method \ding{202} & 0.88 & 0.50 & 0.70 & 0.69 & 0.41 & 0.22 & 0.44 & 0.36 & 0.27 & 0.14 & 0.11 & 0.17 & 0.20 & 0.08 & 0.09 & 0.12 \\
     \qquad Method \ding{203} & 0.83 & 0.44 & 0.67 & 0.64 & 0.57 & 0.34 & 0.31 & 0.40 & 0.35 & 0.21 & 0.08 & 0.21 & 0.22 & 0.13 & 0.04 & 0.13 \\
     \qquad Method \ding{204} & 0.86 & 0.47 & 0.61 & 0.64 & 0.60 & 0.37 & 0.30 & 0.42 & 0.31 & 0.17 & 0.11 & 0.20 & 0.18 & 0.10 & 0.06 & 0.11 \\
      \qquad Method \ding{205} & 0.84 & 0.55 & 0.61 & 0.67 & 0.57 & 0.33 & 0.31 & 0.40 & 0.29 & 0.19 & 0.11 & 0.20 & 0.21 & 0.11 & 0.05 & 0.12 \\
      \rowcolor{red!8}
     \qquad \textbf{+O-Edit+} & 0.89 & \textbf{0.61} &\textbf{ 0.78} & \textbf{0.76} & \textbf{0.81} & \textbf{0.55} & \textbf{0.60} & \textbf{0.65} & \textbf{0.68} & \textbf{0.39 }& \textbf{0.55} & \textbf{0.54} & \textbf{0.61} & \textbf{0.42} & \textbf{0.53} & \textbf{0.52} \\
     \midrule
     \multicolumn{17}{c}{\texttt{\textbf{Llama3-8B}}} \\
     \midrule

     \textbf{MEMIT} & 0.85 & 0.51 & 0.22 & 0.52 & 0.50 & 0.35 & 0.10 & 0.32 & 0.28 & 0.10 & 0.05 & 0.14 & 0.18 & 0.06 & 0.05 & 0.10 \\
     \cmidrule(lr){1-17}
     \qquad Method \ding{202} & 0.74 & 0.33 & 0.58 & 0.32 & 0.32 & 0.11 & 0.51 & 0.31 & 0.24 & 0.08 & 0.34 & 0.22 & 0.13 & 0.07 & 0.18 & 0.12 \\
     \qquad Method \ding{203} & 0.83 & 0.50 & 0.24 & 0.52 & 0.72 & 0.37 & 0.08 & 0.39 & 0.44 & 0.19 & 0.08 & 0.23 & 0.40 & 0.13 & 0.08 & 0.20 \\
     \qquad Method \ding{204} & 0.82 & 0.49 & 0.28 & 0.53 & 0.72 & 0.35 & 0.08 & 0.38 & 0.46 & 0.21 & 0.03 & 0.23 & 0.32 & 0.17 & 0.02 & 0.17 \\
     \qquad Method \ding{205} & 0.77 & 0.41 & 0.44 & 0.54 &0.69 & 0.32 & 0.06 & 0.36 & 0.47 & 0.23 & 0.31 & 0.33 & 0.37 & 0.15 & 0.09 & 0.21 \\
     \rowcolor{red!8}
     \qquad+\textbf{O-Edit+} & \textbf{0.88} & \textbf{0.53} & \textbf{0.76} & \textbf{0.72} & \textbf{0.84} & \textbf{0.51} & \textbf{0.45} & \textbf{0.60} & \textbf{0.81} & \textbf{0.50} & \textbf{0.31} & \textbf{0.54} & \textbf{0.79} & \textbf{0.44} & \textbf{0.28} & \textbf{0.50} \\
     \bottomrule
    \end{tabular}
    }
    \label{canany}
    \vspace{-0.4cm}
\end{table}

\begin{figure}
    \centering
    \includegraphics[width=0.5\textwidth]{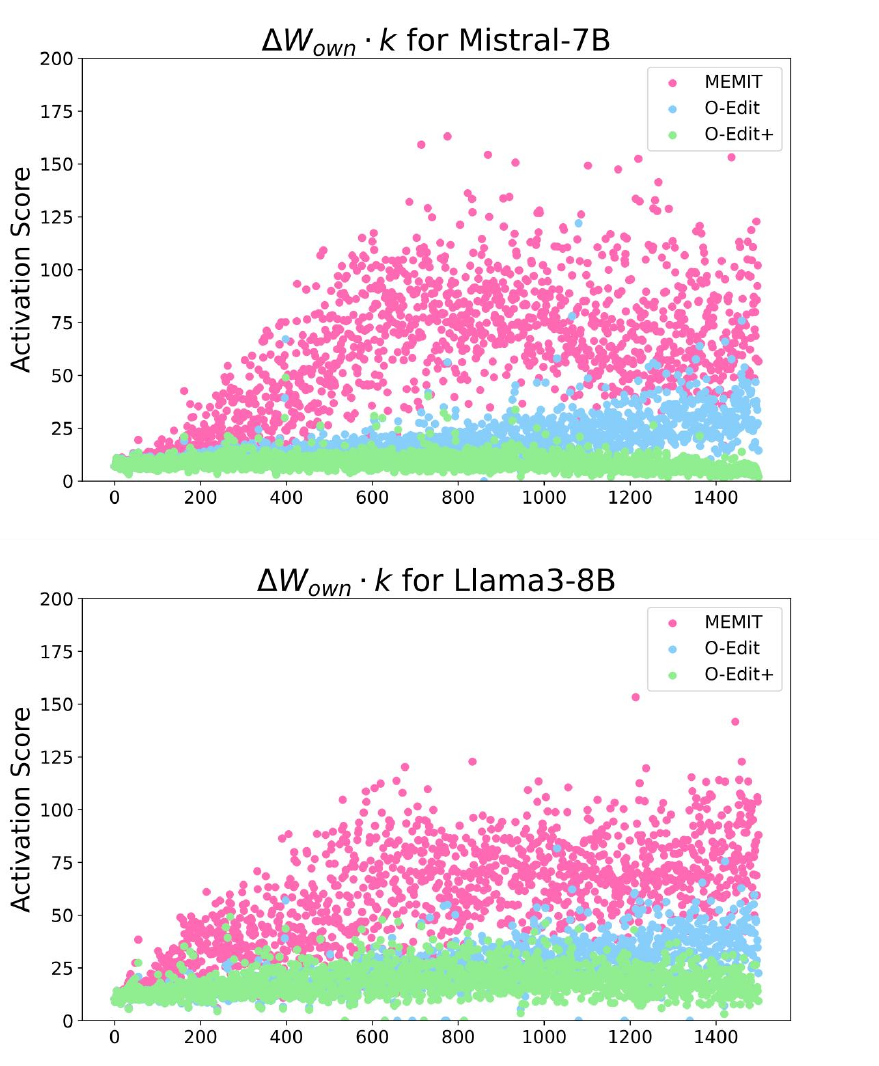}
        \caption{The activation score caused by update parameter itself. \texttt{X-axis}: Num edits.}
        \label{figown}
\end{figure}

\begin{table}[p]
\vspace{-0.5cm}
    \caption{\textbf{Main editing results for ZsRE.} $T$: Num Edits.}
    \centering

    \resizebox{\linewidth}{!}{
    \begin{tabular}{lccc|c|ccc|c|ccc|c|ccc|c}
    \toprule
    \multirow{3}{*}{\textbf{Method}} & \multicolumn{16}{c}{\textbf{COUNTERFACT}} \\
     \cmidrule(lr){2-17}
     & \multicolumn{4}{c|}{$T=200$} & \multicolumn{4}{c|}{$T=500$} & \multicolumn{4}{c|}{$T=1000$} & \multicolumn{4}{c}{$T=1500$} \\
     \midrule
     & Rel. & Gen. & Loc. & Avg. & Rel. & Gen. & Loc. & Avg.  & Rel. & Gen. & Loc. & Avg. & Rel. & Gen. & Loc. & Avg. \\
     \midrule
     \multicolumn{17}{c}{\texttt{\textbf{Mistral-7B}}} \\
     \midrule
    \textbf{ROME} & 0.82 & 0.41 & 0.38 & 0.53 & 0.32 & 0.22 & 0.08 & 0.20 & 0.30 & 0.17 & 0.06 & 0.18 & 0.31 & 0.15 & 0.06 & 0.17 \\
    \cmidrule(lr){1-17}
     \qquad \textbf{+R-Edit} & 0.95 & 0.49 & 0.47 & 0.64 & 0.27 & 0.18 & 0.08 & 0.17 & 0.31 & 0.13 & 0.05 & 0.16 & 0.31 & 0.15 & 0.06 & 0.17 \\
     \qquad \textbf{+WilKE} & 0.88 & 0.44 & 0.43 & 0.58 & 0.41 & 0.27 & 0.10 & 0.26 & 0.27 & 0.17 & 0.06 & 0.17 & 0.29 & 0.19 & 0.05 & 0.17 \\
     \qquad \textbf{+PRUNE} & 0.92 & 0.30 & 0.82 & 0.68 & 0.77 & 0.32 & 0.53 & 0.54 & 0.36 & 0.19 & 0.34 & 0.30 & 0.33 & 0.21 & \textbf{0.27} & 0.27 \\
     \rowcolor{green!8}
     \qquad \textbf{+O-Edit} & 0.99 & 0.42 & 0.73 & 0.71 & 0.77 & 0.41 & 0.51 & 0.49 & 0.45 & 0.18 & 0.29 & 0.31 & 0.35 & 0.20 & 0.20 & 0.25 \\
     \rowcolor{red!8}
     \qquad \textbf{+O-Edit+} & \textbf{0.99} & \textbf{0.46} & \textbf{0.75} & \textbf{0.73} & \textbf{0.80} & \textbf{0.45} & \textbf{0.51} & \textbf{0.60} & \textbf{0.68} & \textbf{0.42} & \textbf{0.32} & \textbf{0.47} & \textbf{0.43} & \textbf{0.16} & 0.25 & \textbf{0.28} \\
     \cmidrule(lr){1-17}
     \textbf{MEMIT} & 0.95 & 0.50 & 0.38 & 0.61 & 0.52 & 0.37 & 0.14 & 0.34 & 0.31 & 0.20 & 0.06 & 0.19 & 0.24 & 0.10 & 0.06 & 0.13 \\
     \cmidrule(lr){1-17}
     \qquad \textbf{+R-Edit} & 0.96 & 0.49 & 0.41 & 0.62 & 0.40 & 0.19 & 0.40 & 0.44 & 0.32 & 0.22 & 0.06 & 0.20 & 0.26 & 0.16 & 0.07 & 0.16 \\
     \qquad \textbf{+WilKE} & \textbf{0.99} & 0.50 & 0.47 & 0.65 & 0.75 & 0.47 & 0.23 & 0.48 & 0.25 & 0.20 & 0.06 & 0.17 & 0.28 & 0.15 & 0.04 &0.16 \\
     \qquad \textbf{+PRUNE} & 0.83 & \textbf{0.53} & 0.47 & 0.61 & 0.76 & \textbf{0.52} & 0.29 & 0.52 & 0.65 & \textbf{0.45} & 0.22 & 0.44 & 0.43 & 0.27 & 0.12 & 0.27 \\
      \rowcolor{green!8}
      \qquad \textbf{+O-Edit} & 0.97 & 0.40 & 0.65 & 0.67 & \textbf{0.88} & 0.42 & 0.43 & 0.57 & \textbf{0.76} & 0.41 & 0.39 & \textbf{0.52} & \textbf{0.61} & \textbf{0.33} & 0.18 & 0.37 \\
      \rowcolor{red!8}
     \qquad \textbf{+O-Edit+} & 0.94 & 0.33 &\textbf{0.80} & \textbf{0.69} & 0.82 & 0.33 & \textbf{0.60} & \textbf{0.58} & 0.69 & 0.31 & \textbf{0.54} & 0.51 & 0.60 & 0.26 & \textbf{0.51} & \textbf{0.43} \\
     \midrule
     \multicolumn{17}{c}{\texttt{\textbf{Llama3-8B}}} \\
     \midrule
     \textbf{ROME} & 0.84 & 0.63 & 0.23 & 0.56 & 0.69 & \textbf{0.62} & 0.03 & 0.44 & 0.73 & \textbf{0.60} & 0.03 & 0.45 & 0.74 & \textbf{0.63} & 0.02 & 0.46 \\
     \cmidrule(lr){1-17}
     \qquad \textbf{+R-Edit} & 0.86 & 0.51 & 0.38 &0.58 & 0.62 & 0.57  & 0.10 & 0.43 & 0.56 & 0.47 & 0.01 & 0.35 & 0.56 & 0.47 & 0.02 & 0.35 \\
     \qquad \textbf{+WilKE} & 0.75 & 0.37 & 0.28 & 0.47 & 0.50 & 0.38 & 0.05 & 0.31 & 0.60 & 0.50 & 0.02 & 0.37 & 0.66 & 0.55 & 0.01 & 0.40 \\
     \qquad \textbf{+PRUNE} & 0.90 & 0.57 & 0.33 & 0.60 & 0.77 & 0.50 & 0.24 & 0.50 & 0.83 & 0.41 & 0.21 & 0.48 & 0.79 & 0.36 & 0.18 & 0.44 \\
     \rowcolor{green!8}
     \qquad \textbf{+O-Edit} & \textbf{0.94} & \textbf{0.66} & 0.51 & \textbf{0.70} & 0.77 & 0.51 & 0.22 & 0.50 & 0.78 & 0.47 & 0.16 & 0.47 & 0.77 & 0.48 & 0.14 & 0.46 \\
     \rowcolor{red!8}
     \qquad \textbf{+O-Edit+} & 0.91 & 0.47 & \textbf{0.55} & 0.52 & \textbf{0.82} & 0.46 & \textbf{0.27} & \textbf{0.52} & \textbf{0.84} & 0.49 & \textbf{0.25} & \textbf{0.53} & \textbf{0.82} & 0.42 & \textbf{0.24} & \textbf{0.49} \\
    \cmidrule(lr){1-17}
     \textbf{MEMIT} & 0.93 & \textbf{0.63} & 0.30 & 0.62 & 0.75 & 0.65 & 0.03 & 0.48 & 0.53 & 0.40 & 0.04 & 0.32 & 0.33 & 0.23 & 0.04 & 0.20  \\
     \cmidrule(lr){1-17}
     \qquad \textbf{+R-Edit} & 0.94 & 0.62 & 0.25 & 0.60 & 0.82 & \textbf{0.69} & 0.10 & 0.53 & 0.65 & 0.55 & 0.06 & 0.42 & 0.52 & 0.41 & 0.03 & 0.32 \\
     \qquad+ \textbf{WilKE} & \textbf{0.98} & 0.42 & \textbf{0.70} & \textbf{0.70} & 0.78 & 0.65 & 0.10 & 0.51 & 0.61 & 0.50 & 0.07 & 0.40 & 0.52 & \textbf{0.42} & 0.05 & 0.33 \\
     \qquad \textbf{+PRUNE} & 0.97 & 0.56 & 0.50 & 0.67 & 0.87 & 0.60 & 0.43 & 0.63 & 0.56 & 0.34 & 0.40 & 0.43 & 0.46 & 0.30 & 0.29 & 0.35 \\
     \rowcolor{green!8}
     \qquad \textbf{+O-Edit} & 0.96 & 0.42 & 0.52 & 0.63 & \textbf{0.90} & 0.49 & 0.41 & \textbf{0.60} & \textbf{0.77} & \textbf{0.51} & 0.32 & \textbf{0.53} & 0.55 & 0.40 & 0.27 & 0.40 \\
     \rowcolor{red!8}
     \qquad+\textbf{O-Edit+} & 0.97 & 0.40 & 0.59 & 0.65 & 0.85 & 0.37 & \textbf{0.46} & 0.56 & 0.73 & 0.32 & \textbf{0.38} & 0.48 & \textbf{0.65} & 0.29 & \textbf{0.36} & \textbf{0.43} \\
     \bottomrule
    \end{tabular}
     }
    \label{zsrezsre}
    \vspace{-0.4cm}
\end{table}

\subsection{Further experiment and discussion}
\label{different}
In this section, we first experiment with the interaction between the \(j\)-th \(k_j\) and \( \Delta W_j \). Then, we analyze the reasons for the smaller value of \( \| \Delta W_\text{unrelated} \cdot k_j \|_2 \) discussed in Section \ref{Further Analysis} regarding O-Edit and O-Edit+. We then compare the similarities and differences between our method and that of \citep{hu2024knowledgesuperpositionunveilingfailures}.

\textbf{Activation on \( \Delta W_\text{own} \)}

In contrast to the second paragraph of Section \ref{Further Analysis}, we aim to further understand the interaction between the \(j\)-th \(k_j\) and \( \Delta W_j \). We calculate the activation score (AS) for each edit as \( \| \Delta W_j \cdot k_j \|_2 \) (\( \|\Delta W_\text{own} \cdot k_j \|_2 \)), as illustrated in Figure \ref{figown}. After 1500 edits, the activation values in MEMIT gradually increase with the number of edits due to the significant activation value \( \|\Delta W_{<j} \cdot k_j \|_2 \) (\( \|\Delta W_\text{unrelated} \cdot k_j \|_2 \)). This phenomenon occurs because completing an edit requires a larger activation value to counteract the influence of previous edits, resulting in a vicious cycle and ultimately poor sequential editing performance. In contrast, the activation values for \( \|\Delta W_{<j} \cdot k_j \|_2 \) (\( \|\Delta W_\text{unrelated} \cdot k_j \|_2 \)) in O-Edit and O-Edit+ remain consistently low, indicating that a large activation value for \( \|\Delta W_\text{own} \cdot k_j \|_2 \) is not necessary to complete a new edit. Consequently, although the activation values are small, O-Edit and O-Edit+ allow for a greater number of effective edits.

\textbf{Theoretical analysis}

Considering MEMIT, we derive from the equations \( \Delta W_{\text{[total]}} = \sum_{i=1}^n \Delta W_i \) and \( \Delta W_\text{unrelated} = \Delta W_{\text{[total]}} - \Delta W_j \) that:
\begin{equation}  
\label{eq29}
    \begin{aligned}  
        \| \Delta W_\text{unrelated} &\cdot k_j \|_2 = \| (\Delta W_{\text{[total]}} - \Delta W_j) \cdot k_j \|_2 \\
        &= \left\| \sum_{\substack{i=1,i \neq j}}^{n} \Delta W_i \cdot k_j \right\|_2  \\
        &= \left\| \sum_{\substack{i=1;i \neq j}}^{n} R_i k_{*;i}^T (\lambda K K^T + k_{*;i} k_{*;i}^T)^{-1} \cdot k_j \right\|_2 \\
        &= \sum_{\substack{i=1;i \neq j}}^{n} \left( R_i k_{*;i}^T (\lambda K K^T + k_{*;i} k_{*;i}^T)^{-1} k_j \right)^T  \sum_{\substack{i=1;i \neq j}}^{n} R_i k_{*;i}^T (\lambda K K^T + k_{*;i} k_{*;i}^T)^{-1} k_j \\
        &= \sum_{\substack{i=1;i \neq j}}^{n} k^T_j \left( (\lambda K K^T + k_{*;i} k_{*;i}^T)^{-1} \right)^T k_{*;i} R^T_i   \sum_{\substack{i=1;i \neq j}}^{n} R_i k_{*;i}^T (K K^T + k_{*;i} k_{*;i}^T)^{-1} k_j
    \end{aligned}  
\end{equation}

Since \( R \) is a column vector, \( R^T \) is a row vector. For any \( R_n \) and \( R_m \) where \( n \neq m \), the updates in O-Edit and O-Edit aim to ensure that each update matrix \( \Delta W \) is orthogonal in the column space, leading to \( R^T_n \cdot R_m \rightarrow 0 \). Consequently, the value of Eqn. \ref{eq29} is smaller than that of MEMIT.

\textbf{Differences and Similarities with \cite{hu2024knowledgesuperpositionunveilingfailures}}

From the perspective of activating \( \| \Delta W_\text{unrelated} \cdot k_j \|_2 \), \citep{hu2024knowledgesuperpositionunveilingfailures} emphasizes the reduction of this metric's activation value through orthogonal row space. They aim to achieve smaller activation values using the expression \( \sum_{\substack{i=1;i \neq j}}^{n} k_{*;i}^T (\lambda K K^T + k_{*;i} k_{*;i}^T)^{-1} k_j \rightarrow 0 \). However, since the variables \( K \) and \( k_* \) are predetermined, their orthogonality cannot be optimized through training methods. To address this, they suggest selecting bottom layers with lower row orthogonality. Yet, this method undermines the extensibility of editing techniques, as knowledge is not solely stored in the lower layers of the model \citep{li2024pmetprecisemodelediting, meng2023locatingeditingfactualassociations, geva2021transformerfeedforwardlayerskeyvalue, geva2022transformerfeedforwardlayersbuild}.

In contrast, O-Edit and O-Edit+ tackle this issue by focusing on orthogonal column space, providing a practical algorithm that supports multiple consecutive edits. These methods can achieve column space orthogonality between update matrices at any layer, effectively reducing \( \| \Delta W_\text{unrelated} \cdot k_j \|_2 \) and facilitating expansion to multi-layer editing.

\end{document}